\documentclass[11pt]{article}
\usepackage{times}
\usepackage{latexsym}
\usepackage[T1]{fontenc}
\usepackage[utf8]{inputenc}
\usepackage{microtype}
\usepackage{inconsolata}
\usepackage[table,dvipsnames]{xcolor}
\usepackage{acl}
\usepackage{multirow}
\usepackage[ruled, lined, linesnumbered, commentsnumbered, longend]{algorithm2e}
\usepackage{algpseudocode}
\usepackage{comment}
\usepackage{caption}
\usepackage{array}
\usepackage{float}
\usepackage{tabularx}
\usepackage{makecell}
\usepackage{mathtools}
\usepackage{amsmath}
\usepackage{graphicx} 
\usepackage{tabularx}
\usepackage{makecell}
\usepackage{todonotes}
\usepackage{multirow}
\usepackage{caption}
\usepackage{balance}
\usepackage{lipsum}
\usepackage{adjustbox}
\usepackage{comment}
\usepackage{subcaption}
\usepackage{booktabs}
\usepackage{color, colortbl}
\usepackage[nolist,nohyperlinks]{acronym} 
\usepackage{tcolorbox}
\usepackage{xcolor}
\newcommand{\new}[1]{\textcolor{red}{#1}}
\newcommand{\swatch}[2]{\colorbox[HTML]{#1}{\strut\,\textbf{#2}\,}}

\usepackage[normalem]{ulem}

\usepackage[english]{babel}
\usepackage{hyperref}
\addto\captionsenglish{%
}
\addto\extrasenglish{%
}

\newtcolorbox{promptbox}[1][]{
  colback=gray!5!white,      
  colframe=gray!75!black,   
  title=\textbf{Prompt Template},
  fonttitle=\bfseries\sffamily,
  coltitle=white,            
  boxrule=0.8pt,             
  rounded corners,           
  arc=3pt,                   
  left=6pt, right=6pt, top=6pt, bottom=6pt,
  #1
}
\newcommand{\placeholder}[1]{\textcolor{blue}{\texttt{\{#1\}}}}
\title{Decoupling Internal Representational Changes and Causal Importance in Fine-Tuned Large Language Models}

\author{
  Lingfang Li$^\dagger$ \quad Procheta Sen$^\dagger$ \quad Shubham Das$^\clubsuit$ \quad Danushka Bollegala$^\dagger$$^\diamond$\\
  University of Liverpool, United Kingdom$^\dagger$ \quad IIEST, Shibpur, India$^\clubsuit$ \quad Amazon$^\diamond$\\
  {\tt \{L.Li85,procheta.sen,danushka\}@liverpool.ac.uk$^\dagger$}\\
  {\tt 2026itm016.shubham@students.iiests.ac.in$^\clubsuit$}
}

\begin{document}
\begin{acronym}[nolist] 
\acro{LLMs}{Large Language Models}
\acro{NLP}{Natural Language Processing}
\acro{KL} {Kullback-Leibler}
\acro{FT} {Fine-Tuning}
\acro{EAP} {Edge Attribution Patching}
\end{acronym}
\maketitle
\begin{abstract}
Fine-tuning has emerged as a widely adopted approach for adapting \ac{LLMs} to a variety of downstream tasks. However, how it reshapes their internal mechanisms remains poorly understood. To address this, we investigate how fine-tuning alters internal representations in LLMs, including attention patterns and layer-wise activations, and examine whether these changes are linked to task-relevant components identified by \ac{EAP} (e.g., attention heads and logit-level activations) that drive task performance. We find that {\ac{EAP}-identified components} are concentrated within specific layers, indicating a degree of functional localisation in how models internalise task-specific behaviour. Notably, the distribution of these components across layers is largely uncorrelated with the layers undergoing the most substantial representational changes during fine-tuning. Furthermore, we observe that overlap in \ac{EAP}-identified components across tasks does not translate into cross-task performance transfer if the tasks are different in nature (e.g., classification vs. generative tasks). More specifically, fine-tuning on one task can lead to a degradation of performance on another when the two tasks exhibit a high degree of overlap in their \ac{EAP}-identified components. Our code is available here. \footnote{\url{https://github.com/LingfangLi/repr-change-vs-causal-importance}}
\end{abstract}

\section{Introduction} \label{sec:intro}
\ac{FT} has been widely used across diverse domains, including high-stake domains like biomedical \citep{wu2024pmcllamabuildingopensourcelanguage,Lee_2019,liu2024moleculargptopenlargelanguage,Luo_2022} and legal applications \citep{chalkidis-etal-2020-legal,colombo2024saullm7bpioneeringlargelanguage,tewari2024legalprobertclassificationlegalprovisions,Shu_2024}, \begin{figure}[htb]
    \centering
    \includegraphics[width=0.85\linewidth]{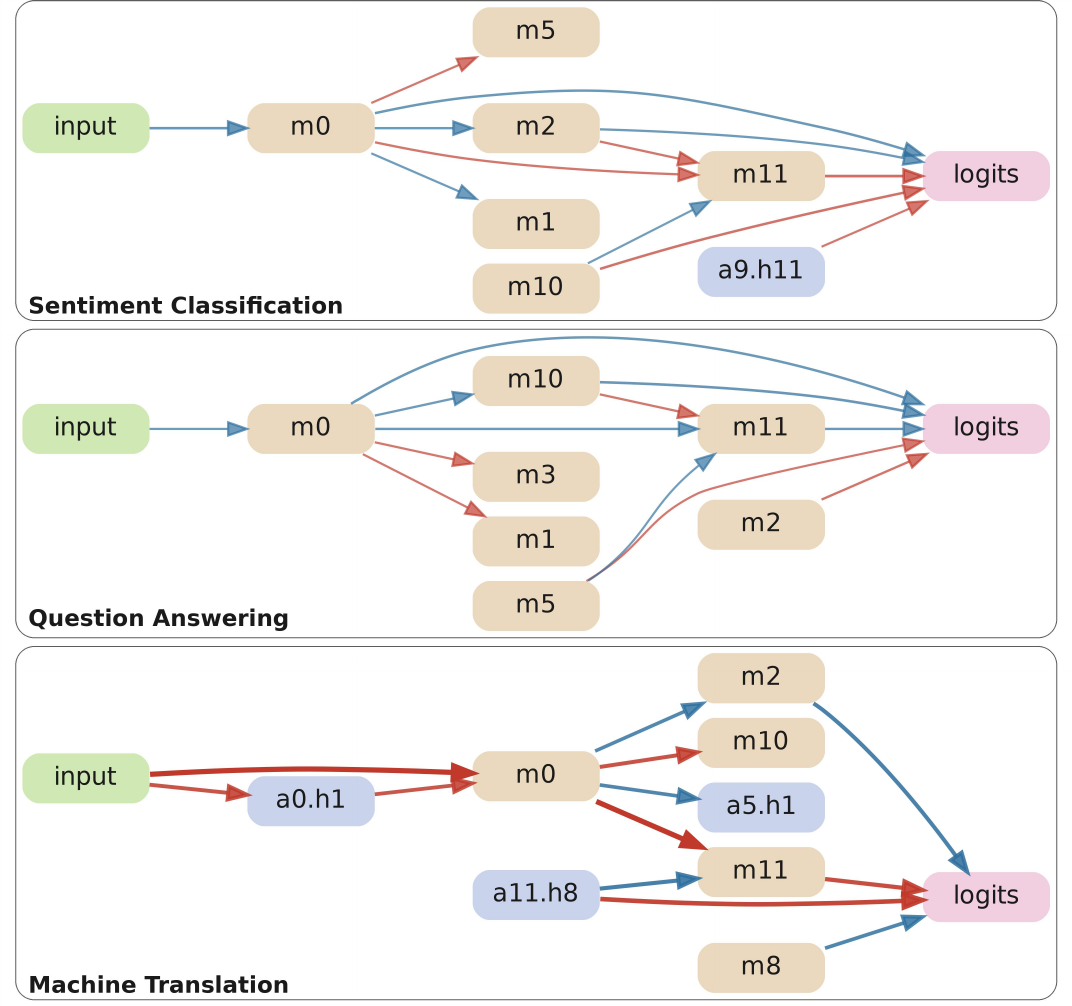}
    \caption{Examples of task-specific \ac{EAP}-identified components (e.g., attention heads, MLPs) 
    for sentiment classification (top), question answering (middle), and machine translation (bottom). Notations are described in \autoref{desc:notation}.} 
    \label{fig:Intro}
\end{figure}    
\citep{jain2023mechanistic,rudman-etal-2023-outlier}. However, there remains a limited understanding of how model-level representational changes relate to the underlying \ac{EAP}-identified components (e.g., embedding nodes, attention heads, MLPs, and logit nodes) that are primarily responsible for task performance. It also remains unclear whether shared \ac{EAP}-identified components across tasks drive cross-task performance transfer, a question that is crucial for developing efficient and principled \ac{FT} strategies.

To address the above-mentioned issue, we study layerwise attention patterns and representation-level changes before and after \ac{FT}. The reason for studying layerwise representations is to understand how task-level information encoded in every layer changes before and after \ac{FT}. We used a causal attribution-based approach from mechanistic interpretability to find the causally relevant components in the fine-tuned model. Our primary objective is to investigate whether the layer undergoing the maximum representational change is also similar to the layer having the highest contribution to the \ac{EAP}-identified components responsible for the performance of the task.

We investigate \ac{FT} across three different \ac{NLP} tasks (a) Sentiment Classification (label prediction), (b) Machine Translation (sequence generation), and (c) Question Answering (span extraction). To increase the generalisability of our findings, for each task, we select two datasets, resulting in six different datasets in total. Our experiments are conducted on GPT-2 Small (124M), Llama-3.2-1B, Qwen2-0.5B, Llama-2-7B. \autoref{fig:Intro} provides an illustrative example of \ac{EAP}-identified components for three tasks (i.e., sentiment classification (SST-2), question answering (SQuAD), and machine translation (KDE4)), visualising a subset of the top-ranked components identified by our analysis. It can be observed that different kinds of tasks have similar and different causally relevant components. For example, MLP5 is present in both sentiment and question answering task, and the first attention head in the fifth layer (a5h1) is present only in machine translation. Broadly, our contributions can be summarised as follows.
\begin{itemize}
     \item We perform a comprehensive causal component analysis rather than focusing only on the attention heads or feedforward layer similar to \citeauthor{yu-ananiadou-2025-locate} in three different NLP tasks (\autoref{sec:localization}). We perform further analysis to quantify the overlap between known functional components, such as induction heads, and the identified causally relevant components in our research scope. In addition, we investigate the relative contribution of different component types (e.g., MLPs, attention heads, and logit nodes) among the top causally relevant components for each task (\autoref{sec:causal}).
     
     \item We show that the layers undergoing the largest changes in attention patterns and representations during \ac{FT} are not the layers that contain the most \ac{EAP}-identified components for task performance ( \autoref{sec:finetuningdynamics} and  \autoref{sec:localization}). 

    \item In contrast, for models fine-tuned on different tasks, we observe no performance gains despite the presence of overlap of \ac{EAP}-identified components. This suggests that differences in task nature (e.g., classification vs. generation) restrict the effective reuse of \ac{EAP}-identified components across tasks (\autoref{sec:generalisability}). 
\end{itemize}

\section{Related Work}\label{sec:related_work}

\paragraph{Mechanistic Studies of Fine-Tuning.}
Existing work has explored several techniques to understand the internal mechanisms of \ac{FT} techniques. Prior studies have examined parameter, feature, and representation-level dynamics. \citet{mukherjee2025reinforcementlearningfinetunessmall} showed that \ac{FT} can match full-update performance while modifying only sparse parameter subsets, \citet{xu2025trackingfeaturedynamicsllm} traced feature evolution across different phases of \ac{FT}, \citet{jain2023mechanistic} used pruning and probing on procedurally defined synthetic tasks and argued that \ac{FT} acts as a lightweight wrapper over pre-trained capabilities. \citet{ren2025learning} proposed a decomposition framework tracing how individual training examples shape predictions. Recent circuit-based work further studies \ac{FT} through mechanistic interpretability. \citet{wang2025towards} tracked circuits across \ac{FT} checkpoints on mathematical tasks and found that \ac{FT} primarily reorganises circuit edges rather than simply adding or removing nodes, motivating a circuit-aware LoRA method for \ac{FT}. \citet{prakash2024fine} showed that fine-tuning can preserve and strengthen existing task mechanisms rather than construct new ones, using entity tracking as a case study. Our work asks a different question: whether the components responsible for task performance are also those that undergo the most substantial changes during fine-tuning, and evaluates this across multiple task families. These works characterise what changes during \ac{FT}, but they do not relate model-level representational change to the \ac{EAP}-identified components responsible for task performance, which is the gap our work addresses. 

\paragraph{Component and Circuit-Level Analyses of LLM Internals.}
A complementary line of work focused on localisation of internal components for different types of model behaviours. Prior studies identified task-relevant neurones in feed-forward networks \citep{geva2021ffn,wang2022skill,xu2025let,pan-etal-2024-finding}, used probing to investigate how task information is distributed across layers \citep{belinkov-2022-probing}, or recover task-relevant circuits through attribution patching and automated circuit discovery \citep{attribution, hanna2024have,conmy2023}. While these methods provide evidence that specific neurones, heads, layers, or edges contribute to model behaviour, many analyses focus on isolated component types (e.g., only on attention heads or only on feed-forward network) or on pre-trained models. 
    \citet{leng2025towards} explored different task-specific neurones, specifically focusing on feed-forward layer networks to develop an efficient \ac{FT} approach. \citet{pmlr-v202-panigrahi23a} showed that a small, localized subset of model parameters can be sufficient to capture task-specific skills through parameter grafting. In contrast to the above-mentioned line of work, we use \ac{EAP} to assign importance scores to edges throughout the model and select the top-\(k\) task-relevant edges across three different types of tasks. We then derive the corresponding sets of model components from these selected edges and examine whether their layer-wise causal importance aligns with the layers that change most during fine-tuning.

\section{Methodology}\label{sec:methodology}
Here, we describe the frameworks used to analyse model-level internal changes and identify \ac{EAP}-identified components.
\subsection{Quantifying Model-Level Changes} 
 We focus on attention patterns and representation-level changes because they capture two complementary aspects of model behaviour under \ac{FT}. Attention patterns provide insights into how the model routes information across tokens, reflecting dependency structure and contextual reasoning \citep{vaswani2017attention,clark-etal-2019-bert,voita-etal-2019-analyzing,NEURIPS2019_2c601ad9}. In contrast, representation-level analyses capture how \ac{FT} alters the geometry of the model’s internal feature space, which encodes semantic and syntactic information \citep{ethayarajh-2019-contextual,hewitt-manning-2019-structural,kovaleva-etal-2019-revealing}. Such changes reflect shifts in the model’s learned abstractions beyond individual components.
 
 \paragraph{Measuring Attention Pattern Changes.} We first compare attention patterns before and after \ac{FT} on each head across all the layers in an LLM. Given a fixed layer $\ell$ and a head $h$, for the $k$-th test instance, tokenised into a sequence of length $d_k$, let $A^{\mathrm{PT},k}_{\ell,h}, A^{\mathrm{FT},k}_{\ell,h} \in [0,1]^{d_k \times d_k}$ denote the corresponding attention pattern in the pre-trained and fine-tuned model, respectively. Each row in $A_{\ell,h,i,:}$ is a softmax output and thus a probability distribution over the $d_k$ key positions ($A_{\ell,h,i,j}=0$ for $j>i$ under causal masking), making the \ac{KL} divergence well defined between corresponding rows. Below, we measure the change magnitude of head $h$ in layer $\ell$ by averaging the row-wise \ac{KL} divergence over all $d_k$ query positions, and then over $n$ test instances as follows:

\begin{equation}
     \Delta A_{\ell,h} =
     \frac{1}{n}\sum_{k=1}^n  \frac{1}{d_k}\sum_{i=1}^{d_k}\operatorname{KL}\!\left(A^{\mathrm{PT},k}_{{\ell},h,i}\,\big\|\,A^{\mathrm{FT},k}_{{\ell},h,i}\right)
     \label{eq:change}
 \end{equation}
Taking the pre-trained pattern as the reference, \autoref{eq:change} measures how much \ac{FT} shifts head $(\ell,h)$. We then average over the $H$ heads, $\Delta A_{\ell}=\frac{1}{H}\sum_{h=1}^{H}\Delta A_{\ell,h}$, to obtain the average change in attention pattern per layer.

\paragraph{Measuring Representation Level Changes.}
\label{sec:logit_lens}
To estimate whether task-relevant information remains encoded within different layers of fine-tuned LLMs in a way similar to that of the pretrained model, similar to \citet{geva-etal-2023-dissecting} and \citet{jiang-etal-2024-large}, we adopt a probing framework using logit lens \citep{nostalgebraist2020logitlens}. Specifically, for each test sample $s$ and layer $\ell$, 
we map the intermediate hidden state representation to the vocabulary space using the same output prediction head used for the final-layer representation, obtaining a per-layer prediction $\hat{y}^{(s)}_{\ell}$ for every sample $s \in \mathcal{S}$. The probing procedure is identical across tasks, what differs is the metric $\mathcal{M}$ used to compare $\hat{y}^{(s)}_{\ell}$ against the ground truth $y_s$.

\begin{equation}
\label{eq:mrr}
\mathcal{M}_\ell \;=\; 
  \frac{1}{|\mathcal{S}|} \sum_{s \in \mathcal{S}} 
  \mathcal{M}\!\bigl(\hat{y}^{(s)}_{\ell},\, y_s\bigr).
\end{equation}

Equation \ref{eq:mrr} describes the mathematical formulation of the metric $\mathcal{M_\ell }$ used to measure the effectiveness of the representation obtained from a layer $\ell$ for an LLM for a task. We instantiate $\mathcal{M_\ell }$ as binary classification accuracy for sentiment classification, BERTScore \citep{zhang2020bertscore} for machine translation, and SQuAD F1 \citep{squad} for question answering. A higher $\mathcal{M}_\ell$ indicates that the representation at layer~$\ell$ encodes more sufficient information to estimate the correct output, allowing us to compare how task-aligned information is distributed across layers in the base and fine-tuned models.

 \subsection{Quantifying Causal Relevance} \label{sec:task_circuit} Although the model can undergo multiple changes during FT, there may be certain components which are potentially causally responsible for the performance of a model in a task. To understand this, similar to \citet{wu2023interpretability}, we used the \ac{EAP} technique \cite{attribution}  to estimate the important task-related components from an LLM. 

 Broadly speaking, \ac{EAP} identifies the importance of each edge in an LLM, where an edge $E_{u \to v}$ exists between a source node $u$ and a destination node $v$ (e.g., outputs of previous layers and inputs of subsequent components). An edge essentially consists of two components while doing analysis related to causal relevance, we consider the union of all the components corresponding to top-k edges identified by \ac{EAP}. \ac{EAP} assigns a score to each edge. The saliency score $S_{u \to v}$ of an edge $E_{u \to v}$ is defined as the expected cumulative causal contribution to the task metric $\mathcal{L}$. To estimate the saliency score of an edge, we essentially corrupt an edge and observe the change in the model output. A first-order Taylor expansion is used to approximate the effect of corruption.  \autoref{eq:EAP} gives the mathematical formulation of the saliency score for an input $x$.
 
\begin{equation}
     S_{u \to v} = \left| E_{x \sim \mathcal{D}} \left[ 
     \left\langle \nabla_{\mathbf{a}_{v}} \mathcal{L}(x), \Delta \mathbf{a}_{v,x} \right\rangle \right] \right|
\label{eq:EAP}
 \end{equation}
where $\Delta \mathbf{a}_{v,x}$ is the change in activation due to applying corrupted input to node u. Mathematically,
 \begin{align}
 \Delta \mathbf{a}_{v,x} = \mathbf{a}_{v}(x) - \mathbf{a}_{v}(x')
 \end{align}
In this formulation, $\mathbf{a}_{v}(x)$ denotes the activations generated by the input
$x$, while $\mathbf{a}_{v}(x')$ denotes those from a corrupted prompt $x'$. The corrupted input $x'$ is specifically constructed to remove key task-related information, thereby allowing us to isolate the causal necessity of each edge. The term $\nabla_{\mathbf{a}_{v}} \mathcal{L}(x)$ represents the sensitivity gradient of the metric with respect to the input of the node $v$ on clean data; $\langle \cdot, \cdot \rangle$ denotes the inner product; $\mathcal{D}$ represents the task-specific dataset; $T$ is the sequence length; $i$ denotes the token position. Similar to \citet{attribution}, we implement the difference in logit values corresponding to ground truth and the predicted output by the \ac{LLMs}, as the primary metric $\mathcal{L}$ to measure the model’s confidence. For specific tasks, this metric is defined as follows. Since EAP relies on a first-order approximation rather than an actual intervention, we validate the identified components with activation patching. Specifically, we replace their clean-run contributions to the child inputs with the corresponding contributions from the corrupted run. Across all models and tasks, this intervention degrades task performance substantially more than intervening on the same number of randomly selected edges. Further details are in Appendix \ref{ap:necessity}.

\textbf{Metric for Binary Sentiment Classification ($\mathcal{L}_{\text{Sent}}$)} Here we measure the difference between the logit of the target label $y^*$ (denoted as $z_{y^*}$) and that of the highest-scoring incorrect label for a particular input $x$. If certain edges are important, then corrupting them should not only alter the model’s output significantly but also reduce this logit difference. Mathematically $\mathcal{L}_{\text{Sent}}$ is defined in \autoref{sentiment}.

\begin{equation}
    \mathcal{L}_{\text{Sent}} = z_{y^*} - \max_{j \neq y^*} z_j \quad 
    \label{sentiment}
\end{equation}
where $j$ ranges over all class labels.

\textbf{Metric for Generative Tasks ($\mathcal{L}_{\text{QA, MT}}$)}
For generative tasks, token sequences are produced under a teacher-forcing regime. At each decoding step, we compute the logit difference between the ground-truth token ($GT_i$) and the highest-scoring incorrect token within the top-$K$ predicted candidates. This logit difference is then aggregated across all n decoding steps, where n denotes the total number of tokens in the ground-truth sequence ($GT$). \autoref{eq:mt} describes the mathematical formulation for this metric.
\begin{equation}
    \mathcal{L}_{\text{QA, MT}}
    =
    \sum_{i=1}^{|GT|}
    \left(
        z_{GT_i}^{(i)}
        -
        \max_{j \in K_i \setminus \{GT_i\}} z_j^{(i)}
    \right)
    \label{eq:mt}
\end{equation}
where $j$ ranges over the top-$K$ candidate tokens at that step excluding the ground-truth token.

Across all generative task datasets, the output sequence lengths exhibit low variability, with an average standard deviation of 2.12. This indicates that token length variation is minimal and unlikely to introduce any bias in the evaluation metric $\mathcal{L}_{\text{QA, MT}}$.

\section{Experiment Setup}\label{sec:setup}
\paragraph{Dataset Description.}
Existing research \cite{nlpresearch1,nlpresearch2} showed that the three tasks mentioned above are among the most widely used \ac{NLP} tasks. Hence, we focused particularly on these three tasks in our research scope. For \textbf{Sentiment Classification} task, we used two datasets:  Large Yelp Review (Yelp) Dataset \cite{yelp_polarity} and Stanford Sentiment Treebank (SST-2) \cite{sst2}. For \textbf{Question Answering} task, we used two benchmark reading-comprehension-style datasets: the Stanford Question Answering Dataset (SQuAD) v1.1 \cite{squad} and Conversational Question Answering (CoQA) \cite{coqa}. For \textbf{Machine Translation} task, we used KDE4 dataset \cite{tiedemann-2012-parallel} and Tatoeba dataset \cite{artetxe-schwenk-2019-massively}. The details of the datasets are given in \autoref{ap:dataset}.

\begin{figure*}[h!]
    \centering
             \begin{subfigure}[b]{0.45\textwidth}
          \includegraphics[width=\linewidth]{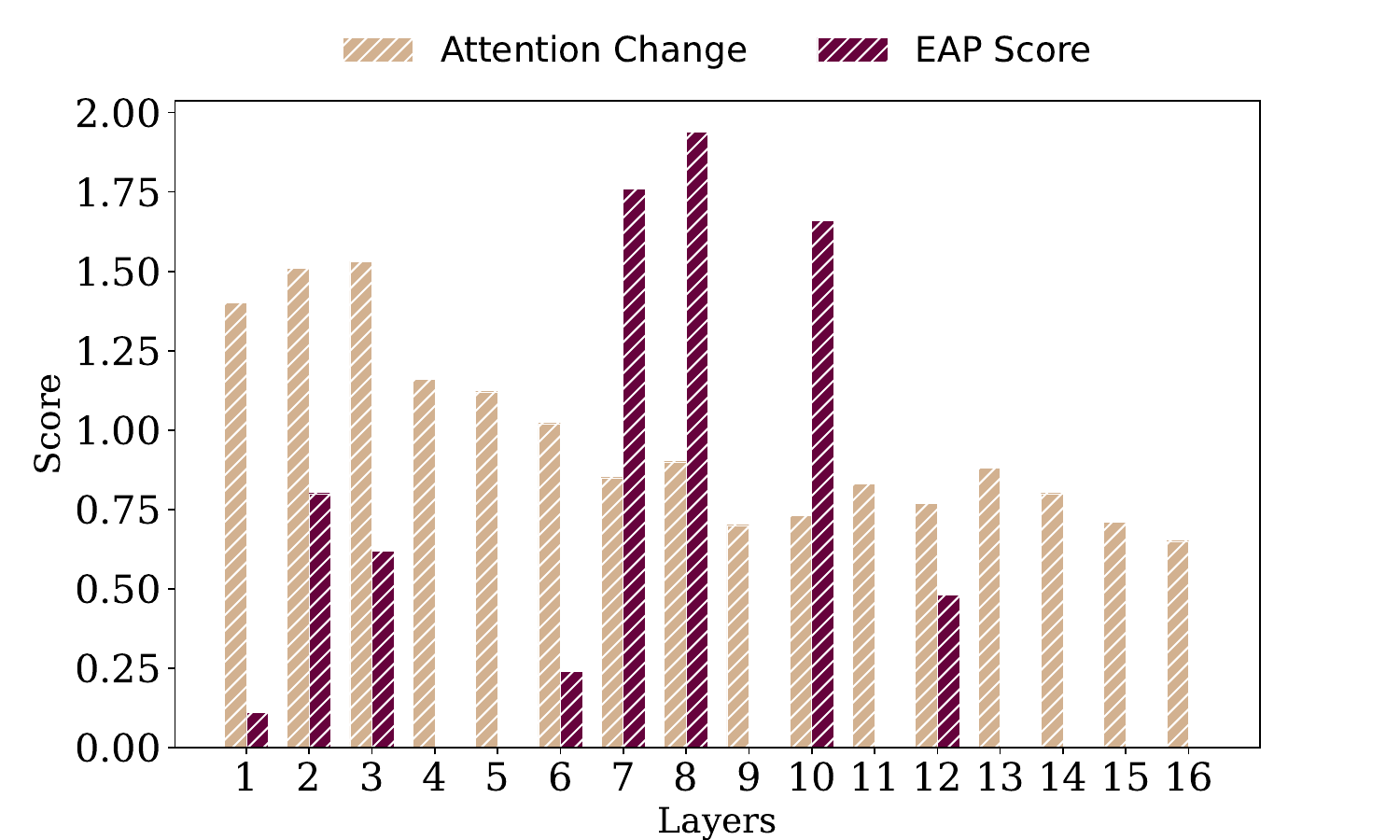}
    \caption{Llama3.2 on SQuAD ($\tilde{H}_{\text{attn}}= 1.00$, $\tilde{H}_{\text{EAP}}=0.59$)} 
   \end{subfigure}
                \begin{subfigure}[b]{0.45\textwidth}
          \includegraphics[width=\linewidth]{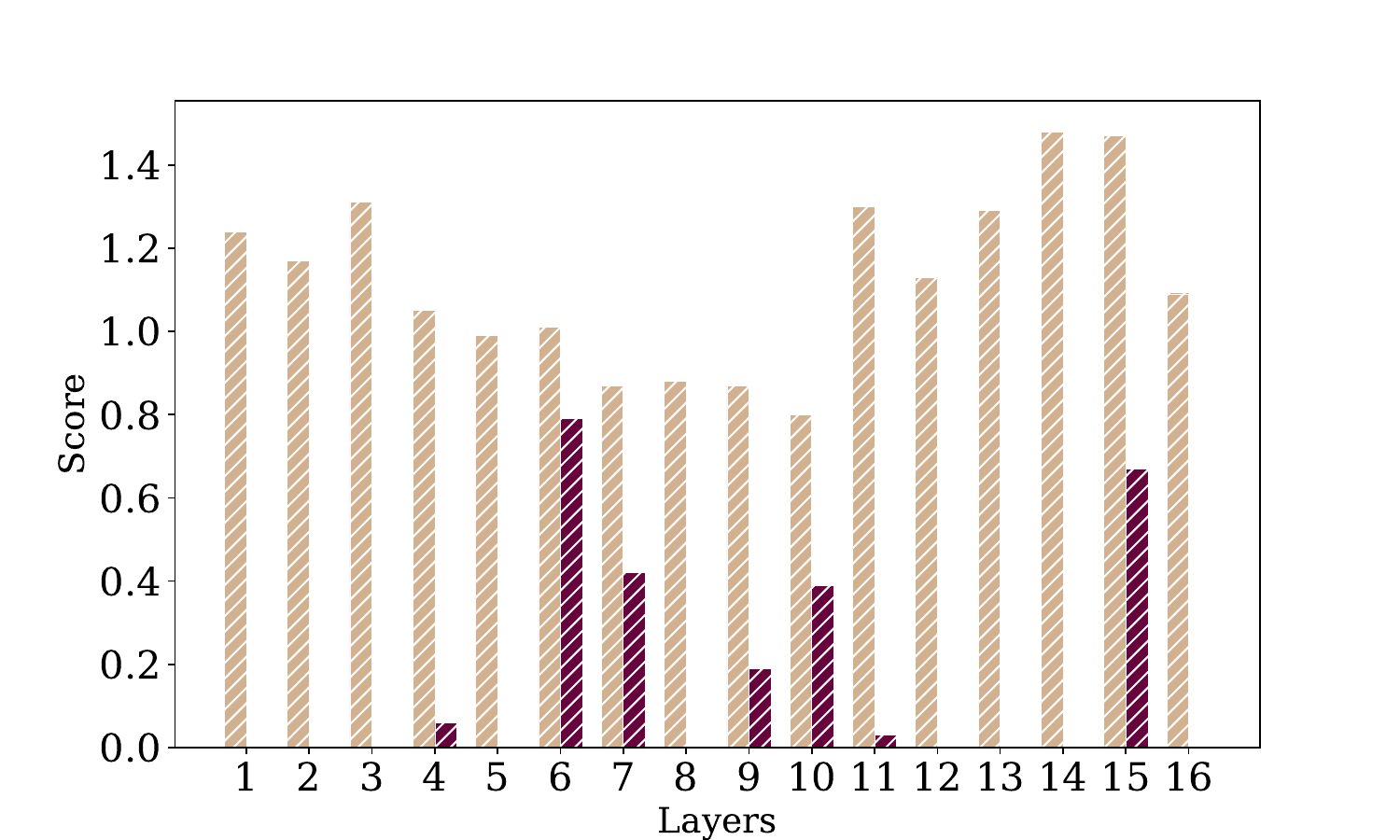}
    \caption{Llama3.2 on KDE4 ($\tilde{H}_{\text{attn}}=0.99$, $\tilde{H}_{\text{EAP}}=0.59$)} 
   \end{subfigure}
      \begin{subfigure}[b]{0.45\textwidth}
          \includegraphics[width=\linewidth]{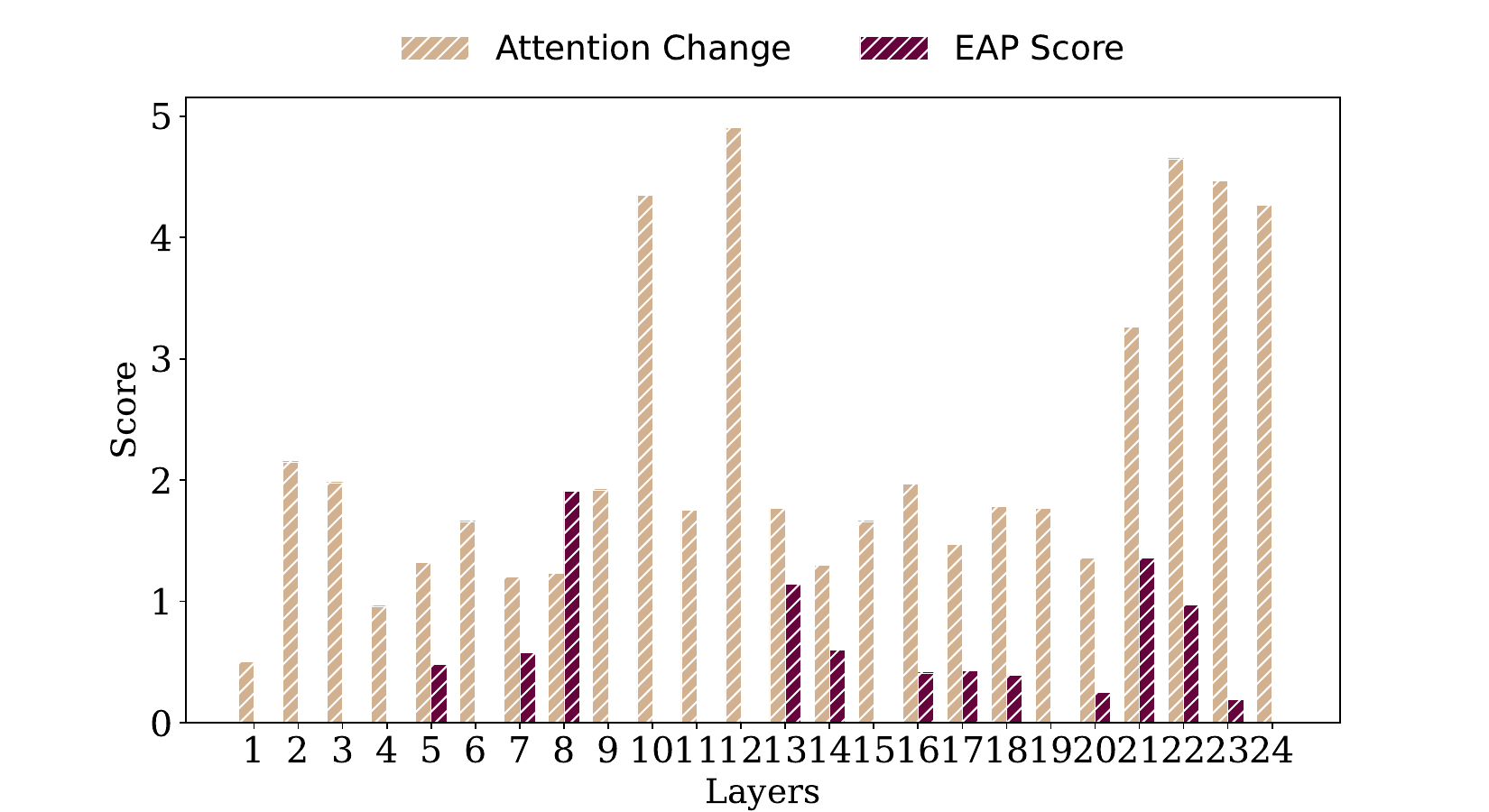}
    \caption{Qwen on SQuAD ($\tilde{H}_{\text{attn}}=0.96$, $\tilde{H}_{\text{EAP}}=0.74$)} 
   \end{subfigure}
     \begin{subfigure}[b]{0.48\textwidth}
          \includegraphics[width=\linewidth]{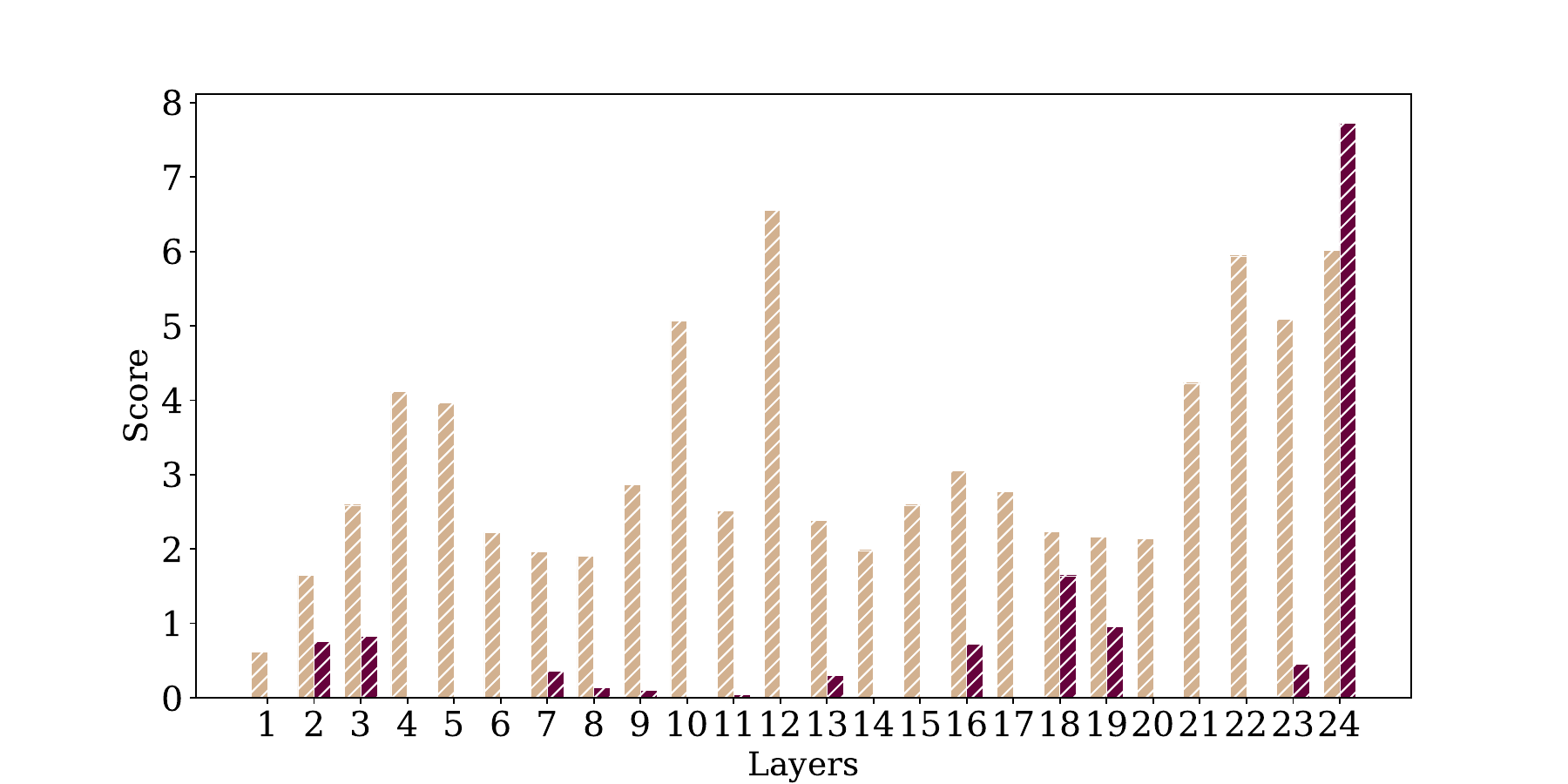}
    \caption{Qwen on KDE4 ($\tilde{H}_{\text{attn}}=0.95$, $\tilde{H}_{\text{EAP}}=0.72$)} 
   \end{subfigure}
   \caption{Layer-wise distribution of average \ac{KL} divergence in the attention matrices before and after \ac{FT}, and summed \ac{EAP} scores across  Fine-tuned Qwen2 and llama-3.2-1B. For GPT-2 Small and  Llama-2-7B refer to \autoref{fig:gpt2-attn-eap} in \autoref{app:rest-results}; results for Qwen2 and Llama-3.2-1B on the selected remaining datasets are shown in \autoref{fig:attn-eap} in \autoref{app:rest-results}. Values in parentheses are the normalised layer-wise entropy
$(\tilde{H}_{\text{attn}},\,\tilde{H}_{\text{EAP}})$; the consistently lower $\tilde{H}_{\text{EAP}}$ indicates that \ac{EAP} importance is more localised across layers. See \autoref{app:entropy} for the entropy definition.}
    \label{fig:eap_attention}
\end{figure*}

\paragraph{Preprocessing.} \label{sc:preprocess}
As described in \autoref{sec:task_circuit}, the \ac{EAP} algorithm used to estimate the \ac{EAP} importance of different components requires corrupted inputs corresponding to each sample input. To generate the corrupted sentences for the sentiment classification task, we masked the tokens most indicative of sentiment in each sentence using the `<mask>' token. These sentiment-bearing tokens were automatically identified using the Meta-Llama-3-8B-Instruct model. Similarly, for the question answering task, the answer span within the context passage was replaced with the `<mask>' token. For the machine translation task, where most source tokens contribute significantly to the output, we replaced all tokens in the English input with `<mask>' to simulate maximal information loss. Examples of generated corrupted data are shown in \autoref{app:corrupted_data}. Although the corrupted token strategies differ across tasks, they follow the same principle of removing information directly relevant to the task output. We do not assume that the resulting corruption severity is identical across tasks; instead, EAP is applied within each task to rank task-relevant components, and our cross-task analyses compare these identified component sets. We also test sensitivity to the corruption token in Appendix \ref{app:corruption token}. Replacing '<mask>' with several alternative tokens produces strongly correlated edge rankings (\(\rho=0.71\text{--}0.72\)), suggesting that the identified components are not specific to the '<mask>' token.

For \ac{EAP} algorithm described in \autoref{sec:task_circuit}, we considered only the components within top $400$  edges for our research scope since we found that most of the important components across all of the \ac{LLMs} and for all the tasks lie within this range. \autoref{fig:ablation-top-edges} in Appendix \ref{ap:ablation} shows the contribution of these components  across different tasks. It can be observed from \autoref{fig:ablation-top-edges} that using only components from top-400 edges preserves at least 90\% of the original task performance across the evaluated settings. 

\paragraph{\ac{FT} Setup.} We used GPT-2 Small, Llama-3.2-1B, Qwen2-0.5B and Llama-2-7B model in our experiment. All four models were fine-tuned in a full \ac{FT} setup, i.e., all model parameters were updated. In \autoref{ap:hyperparameter}, we described in detail other parameters like batch size, epoch used in \ac{FT}.

\section{Results} \label{sec:results}
We present results along three dimensions, namely, (i) the distribution of changes induced by \ac{FT}, (ii) the distribution of causally relevant components after \ac{FT}, and (iii) the relationship between overlap of \ac{EAP}-identified components and the cross-task performance transfer capability of the fine-tuned model. 

\begin{figure*} [h!] 
\includegraphics[width=\textwidth]{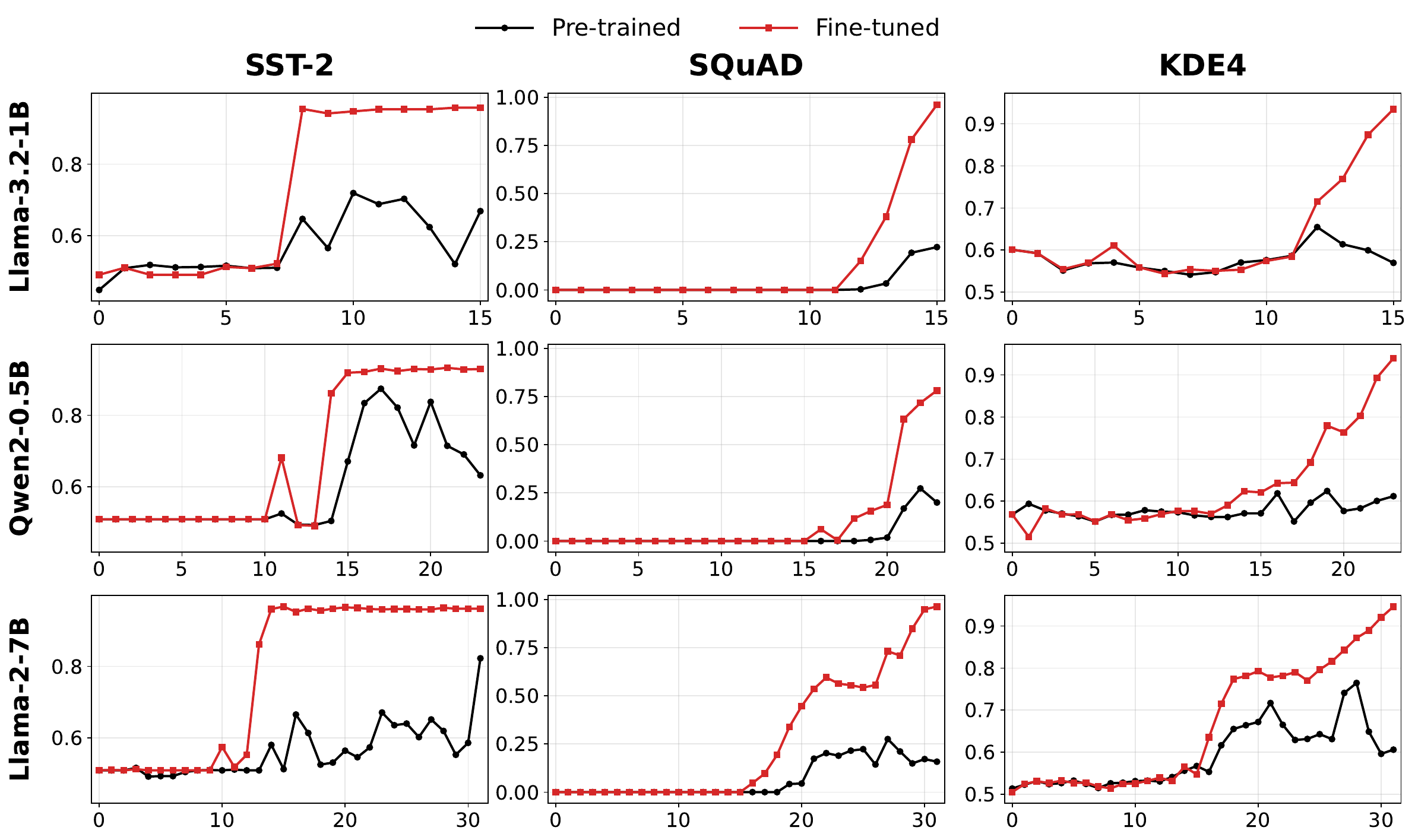}
\caption{Probing results of pretrained model vs. fine-tuned model across all the layers of  Llama-3.2-1B, Qwen2-0.5B, Llama-2-7B for Sentiment Classification (SST-2), Question Answering (SQuAD), and Machine Translation (KDE4). Probing performance for each task is measured by the corresponding task metric as described in Section \ref{sec:logit_lens}. Higher values indicate that the correct task output is more readily recoverable from that layer's representation via the logit lens. The complete layer-wise probing results for all four models across all six datasets are shown in \autoref{fig:gpt2-probing} in \autoref{app:rest-results}. (X-axis: layer index, Y-axis: binary accuracy for SST-2, SQuAD F1 Score for SQuAD, BERTScore for KDE4).}
\label{fig:probing}
\end{figure*}

\subsection{Heterogeneity of Fine-tuning Dynamics} \label{sec:finetuningdynamics}
To demonstrate the non-uniformity of \ac{FT} dynamics, \autoref{fig:eap_attention} illustrates the changes in attention patterns across layers (computed using \autoref{eq:change}), while \autoref{fig:probing} shows the variation in the amount of information encoded in each layer after \ac{FT}.

Two key observations emerge from \autoref{fig:eap_attention}. First, attention pattern change is specific to a particular task and model, and it is not uniform. We quantify this non-uniformity using the normalised layer-wise entropy $\tilde{H}\in[0,1]$ of each distribution (defined in \autoref{app:entropy}). The different values of attention entropy across different variations (i.e., different models and tasks) also further confirm this observation. Secondly, we observe that smaller models exhibit a substantially greater magnitude of representational change during \ac{FT} compared to larger models. For instance, the average KL-divergence values in Llama-3.2-1B span a higher range (approximately 7–10), whereas for Llama-2-7B they remain significantly lower (approximately 0–1). This observation is broadly consistent with prior work on representation preservation and parameter-efficient adaptation \cite{aghajanyan2021better}, which suggests that larger pretrained models possess richer reusable features and therefore require comparatively smaller representational modifications during downstream \ac{FT}. For the rest of the models and the datasets, refer to \autoref{fig:gpt2-attn-eap}  and \autoref{fig:attn-eap} in \autoref{app:rest-results}.

\autoref{fig:probing} shows how task-related information encoded in different layers change after \ac{FT}. The main observations from it are as follows. Firstly, both for pretrained and finetuned setup for all the tasks and for all the models, the probing performance of the initial layers is worse than later layers. Consequently, it can be said that the initial layer representations do not contain much information about the task. This observation is similar to \citet{zhao2024layer} who show that in general, initial layers of LLMs store core language capabilities and the later layer has more derived properties. Secondly, the probing performance in the \ac{FT} setup only changes at later layers mostly. This observation concludes that \ac{FT} doesn't affect the core capabilities of the model. Thirdly, it can be observed clearly that for all the models and for all the tasks the probing performance has increased after \ac{FT}. It indicates that \ac{FT} increases the amount of task-level information encoded in each layer. For GPT-2 Small model and the rest of the datasets, refer to \autoref{fig:gpt2-probing} in \autoref{app:rest-results}. Since probing results are output-mediated and capture layer-wise decodability, we further investigate a direct measure of representational geometry change in each layer rather than passing through the output head of model in \autoref{app:cka}. It shows that across all models and tasks, the most substantial layer-wise representational changes occur in the later layers (in the order of 0.7 to 0.9), specifically from approximately the 90th percentile layer to the final layer.

Overall, from both \autoref{fig:eap_attention}, \autoref{fig:probing} and \autoref{app:cka}, it is evident that the internal change in the model is not uniform across all the layers and it is valid across all the models and tasks. 
\begin{figure*} [h!]  
\begin{subfigure}[b]{0.30\textwidth}
        \includegraphics[width=\linewidth]{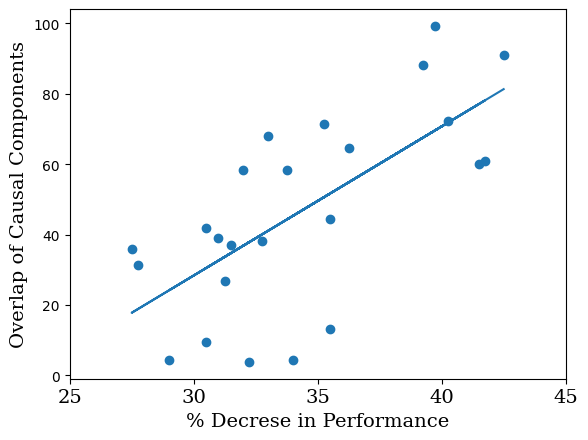}
    \caption{Qwen2, $\rho=0.7, p=0.0002$ }  
    \end{subfigure}
\begin{subfigure}[b]{0.30\textwidth}
        \includegraphics[width=\linewidth]{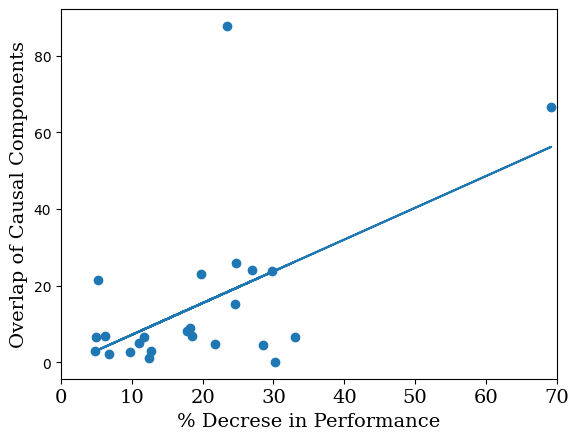}
    \caption{Llama-3.2-1B, $\rho=0.6, p=0.005$}  
    \end{subfigure}
    \begin{subfigure}[b]{0.305\textwidth}
        \includegraphics[width=\linewidth]{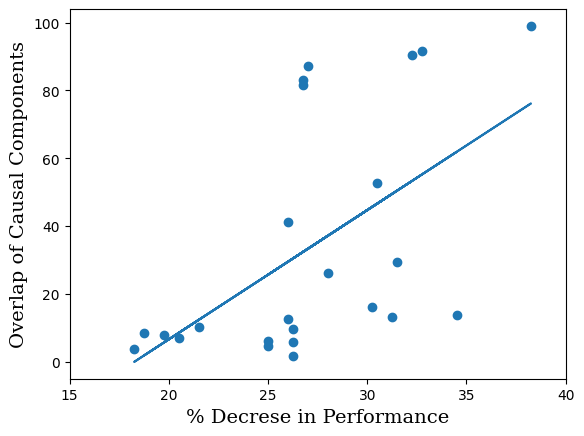}
    \caption{Llama-2-7B $\rho=0.6, p=0.005$}  
    \end{subfigure}
\caption{Correlation between Cross-task Performance and Important Component Overlap for  Llama-3.2-1B, Qwen2-0.5B and Llama-2-7B. $\rho$ is correlation coefficient and p-values show the significance of the correlation coefficient. The result for GPT-2 Small is provided in \autoref{fig:gpt2-correlation}, \autoref{app:rest-results}.}\label{fig:performance_overlap}
\end{figure*}

\subsection{Localisation of \ac{EAP}-Identified Components}\label{sec:localization}
\autoref{fig:eap_attention} presents the layer-wise distribution of the total importance of causally relevant components, where importance scores are computed using \ac{EAP} equation described in \autoref{eq:EAP}. \autoref{eq:EAP} provides score to each edge and then for each layer we show the sum of the scores corresponding to the top-400 edges belonging to that layer.

Two key observations emerge from the EAP score distribution in \autoref{fig:eap_attention}. First, a substantial number of layers exhibit negligible or no contribution to the set of causally relevant components. The entropy score of the EAP score distribution further supports this observation. In contrast to the attention change, the \ac{EAP}-score distribution has a lower entropy ($\tilde{H}_{\text{EAP}}=0.40$--$0.74$; see \autoref{app:entropy}). This pattern is consistent across all models and tasks, suggesting that \ac{EAP}-identified components are highly localised.
Second, the layers that undergo the largest changes in attention patterns do not necessarily correspond to those with the highest \ac{EAP} importance. This indicates that the magnitude of internal change is only weakly correlated with causal relevance.

\subsection{Cross-task Performance Transfer} \label{sec:generalisability}
Here, we analyse whether \ac{FT} a model on Task 1 leads to improvements or degradation in performance on Task 2, where the two tasks are substantially different in nature (e.g., sentiment analysis and machine translation). More specifically, our objective is to examine the relationship between overlap of \ac{EAP}-identified components and cross-task performance transfer. In particular, we investigate whether the overlap in \ac{EAP}-identified components between Task 1 and Task 2 influences performance on Task 2 when the model is fine-tuned on Task 1. To this end, \autoref{fig:performance_overlap} shows the correlation between a decrease in performance and overlap of top $400$ \ac{EAP}-identified components, and for each task, we consider four distinct tasks to compute both performance changes and overlap of \ac{EAP}-identified components. In total, this results in $24$ task pairs for comparison. The list of tasks is in the Appendix, \autoref{tab:performance_overlap} "Dissimilar Task Analysis".

Two main observations emerge from \autoref{fig:performance_overlap}.
Firstly, we observe that \ac{FT} a model on Task 1 generally leads to a degradation in performance on Task 2, relative to the pretrained model’s performance on Task 2. Secondly, we observe a positive correlation across all models (Qwen2, Llama-3.2-1B, and Llama-2-7B) between the degree of overlap of \ac{EAP}-identified components and the extent of performance degradation. See result for GPT-2 Small in \autoref{fig:gpt2-correlation}, \autoref{app:rest-results}. This suggests that higher overlap between the \ac{EAP}-identified components of Task 1 and Task 2 is associated with greater performance decline on Task 2 when the model is fine-tuned on Task 1 and vice versa. A plausible explanation for this phenomenon is that \ac{FT} adapts a specific set of components to the requirements of Task 1, which is unlikely to generalise well to Task 2 when the two tasks are different in nature. To validate that the shared components are responsible for the degradation directly rather than correlational, we perform an intervention in \autoref{app:validation-correlation}: we identify the components that are shared between the two tasks, freeze them, and fine-tune only the remaining parameters on Task 1. Protecting the overlapping components during fine-tuning recovers part of the lost performance on Task 2. So protecting them provides intervention-based evidence that the overlapping components contribute to the observed interference. 

\begin{figure*}[htb]
     \centering
\includegraphics[width=0.7\linewidth]{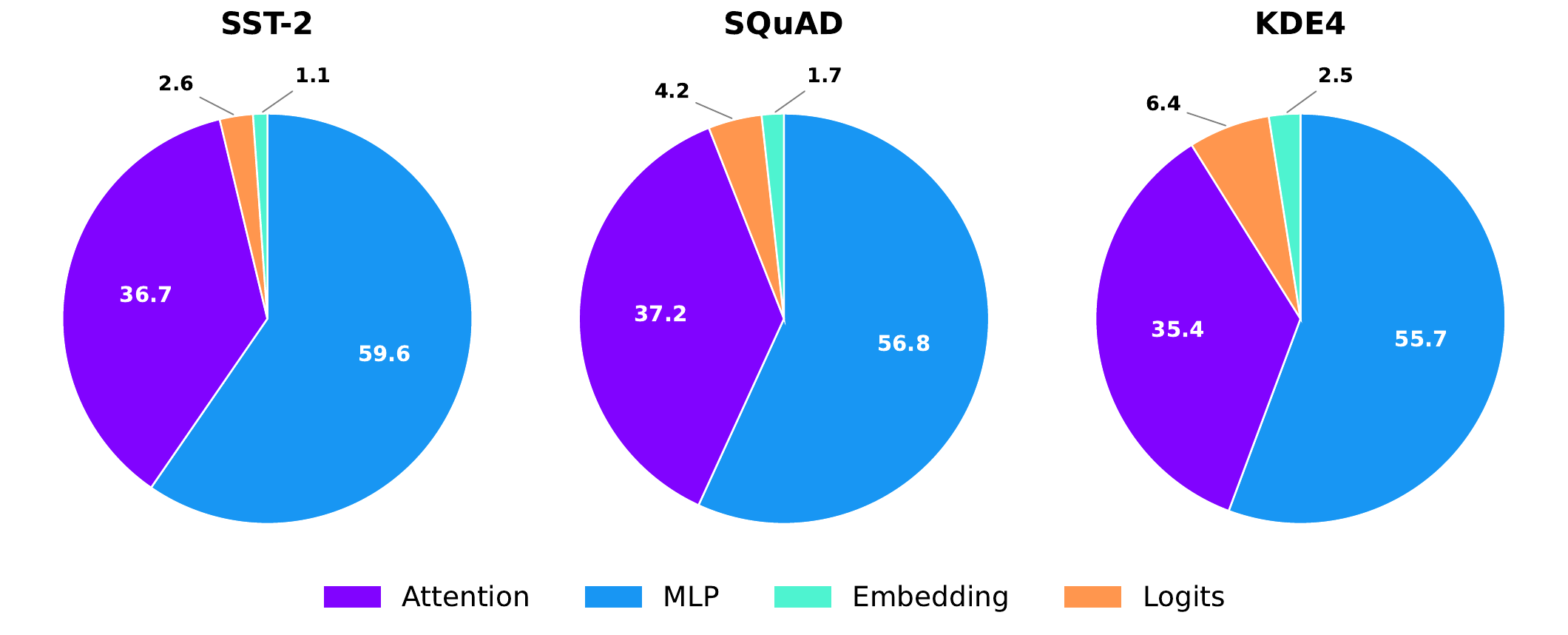}
     \caption{Distribution percentage of important components among the top-400 edges in Llama-3.2-1B across three tasks: Sentiment Classification (SST-2),  Question Answering (SQuAD). Machine Translation (KDE4). 
     The full results for four models and six datasets are given in \autoref{fig:all component distribution} in \autoref{ap:fig:piechart}. 
     }
     \label{fig:component distribution}
 \end{figure*}
\section{Causal Structure Analysis} \label{sec:causal}
To further investigate the important \ac{EAP}-derived components, we analyse the distribution of different architectural elements, specifically embedding nodes, attention heads, MLP neurones, logit nodes, and among the EAP-derived components selected for each task. \autoref{fig:component distribution} presents pie charts illustrating the composition across three tasks in Llama-3.2-1B. More analysis on the rest of the models and datasets are shown in \autoref{fig:all component distribution} in \autoref{ap:fig:piechart}. The results show a consistent pattern across tasks and models: the components spanned by the top-400 EAP-ranked edges are predominantly composed of MLP neurones and attention heads, with relatively limited contribution from other component types. To avoid the frequency bias from the component distribution of the original model, we also provided normalised frequency of different components in Appendix \ref{norm:freq}. It can be seen that the presence of MLPs and logits are multiplied in the normalised frequency.

Given the prominence of attention heads within these important components, we further investigated the specific involvement of \textit{induction heads} \citep{olsson2022}, components responsible for pattern matching within input context. We investigated induction heads following the synthetic repeated-sequence protocol of \citet{olsson2022} (implemented via TransformerLens, \citealp{nanda2022transformerlens}). \autoref{tab:induction_overlap} shows  the Jaccard overlap between the detected induction heads and the task-specific causally important attention heads, computed as the size of their intersection divided by the size of their union. The detection details and the full metric table are provided in \autoref{app:induction}.

We observe three key trends from \autoref{tab:induction_overlap}.  
Firstly, three of the four models exhibit their highest overlap on question answering, with CoQA yielding the maximum overlap for GPT-2 Small (16.66\%), Llama-3.2-1B (6.16\%), and Llama-2-7B (4.34\%). Qwen2-0.5B is the exception, reaching its highest overlap on KDE4 (7.19\%). 
Secondly, the results suggest a task-dependent pattern in the reuse of induction heads. Question answering tends to show relatively higher overlap with induction heads for GPT-2 Small, Llama-3.2-1B and Llama-2-7B. This may be related to the reading-comprehension nature of the QA datasets used in our study, where answers must be identified from or grounded in the input context. Such behaviour is conceptually similar to the pattern-matching mechanisms associated with induction heads \citep{olsson2022}. 
Thirdly, this tendency is not consistent across all models and datasets. Qwen2-0.5B exhibits its highest overlap on KDE4, while its overlap on Tatoeba remains considerably lower. Likewise, MT shows relatively low overlap for the two Llama models. These results suggest that induction-head involvement is not determined solely by task family, but may also depend on model- and dataset-specific factors.

\begin{table}[t]
\centering
\resizebox{0.5\textwidth}{!}{%
\begin{tabular}{llccc}
\toprule 
\textbf{Model} & \textbf{Task} & \textbf{\#Induction Heads} & \textbf{\#Attention Heads} & \textbf{Overlap(\%)} \\
\midrule 


\multirow{6}{*}{\textbf{GPT-2 Small}} 
& Sentiment (Yelp) & 26 & 101 &  8.54\%  \\
& Sentiment (SST-2) & 24 & 80 &  1.96\%\\
& QA (SQuAD) & 24 & 98 &  12.96\%  \\
& QA (CoQA) & 24 & 109 & \underline{\textbf{16.66\%}} \\
& MT (KDE4) & 24 & 99 &  7.89\%\\
& MT (Tatoeba) & 24 & 93 & 7.33\%\\
\midrule 

\multirow{6}{*}{\textbf{Llama-3.2-1B}} 
 & Sent (Yelp) & 62 & 151 & 3.39\% \\
 & Sent (SST-2) & 58 & 146 & 1.49\% \\
 & QA (SQuAD) & 62 & 158 & 5.76\% \\
 & QA (CoQA) & 62 & 162 & \textbf{6.16\%} \\
 & MT (KDE4) & 64 & 167 & 2.21\% \\
 & MT (Tatoeba) & 57 & 132 & 2.71\%\\
\midrule 

\multirow{6}{*}{\textbf{Qwen2-0.5B}} 
 & Sent (Yelp) & 40 & 112 &   2.70\%\\
 & Sent (SST-2) & 35 & 83 & 1.85\% \\
 & QA (SQuAD) & 36 & 97 & 3.10\% \\
 & QA (CoQA) & 39 & 110 & 3.75\%\\
 & MT (KDE4) & 33 & 116 & \textbf{7.19\%}  \\
 & MT (Tatoeba) & 40 & 75 & 1.76\% \\
\midrule 

\multirow{6}{*}{\textbf{Llama-2-7B}} 
 & Sent (Yelp) & 53 & 263 & 1.28\% \\
 & Sent (SST-2) & 52 & 94 &  0.00\% \\
 & QA (SQuAD) & 15 & 270 & 3.70\% \\
 & QA (CoQA) & 61 & 227 & \textbf{4.34\%}  \\
 & MT (KDE4) & 58 & 166 & 0.00\%\\
 & MT (Tatoeba) & 49 & 88 & 0.00\% \\
 
\bottomrule 
\end{tabular}%
}
\caption{Overlap of Induction head vs. Attention heads derived from the top-400 EAP-ranked edges across all models for Sentiment Analysis (Sent), Question Answering (QA) and Machine Translation (MT). For each model the highest overlap is boldfaced and the highest overlap percentage across all the models is underlined.}
\label{tab:induction_overlap}
\end{table}

\section{Conclusion} We studied how \ac{FT} reshapes LLM internals along two aspects: the relationship between representational change and the \ac{EAP}-identified components driving task performance, and the effect of shared \ac{EAP}-identified components on cross-task transfer. Across four models and six datasets, we find that the layers undergoing the largest representational and attention changes do not necessarily coincide with those carrying the highest \ac{EAP} importance. More generally, the magnitude of internal change is only weakly correlated with causal relevance. On cross-task transfer, we find that a high overlap in \ac{EAP}-identified components between two tasks does not aid generalisation; instead, it is associated with greater performance degradation on the second task,
which suggests that components adapted to one task do not transfer benignly to a substantially different one. Together, these results indicate that effective fine-tuning is governed by a small, localised set of \ac{EAP}-identified components rather than by broad representational shifts. 

 In the future, we plan to extend this work to a larger number of tasks. Our findings related to different aspects of localisation and generalisation can also be used to design efficient and effective strategies of \ac{FT} in \ac{NLP}. 
\section*{Limitations}
We restrict our analysis to three \ac{NLP} task families (sentiment classification, question answering, and machine translation). Although these tasks encompass diverse input–output structures, they do not represent the full spectrum of real-world applications, such as reasoning-intensive or multimodal tasks, which limits the transferability of our findings. We also examined only GPT-2 Small and Llama-3.2-1B, Llama-2-7B and Qwen2-0.5B due to the shortage of computing resources, limiting generalisability to larger models that may exhibit different \ac{FT} behaviours. Further work is needed to investigate more models. 
\section*{Ethical Considerations}
This work involves non-sensitive publicly accessible datasets. This work on mechanistic interpretability presents a dual-use dilemma: while understanding \ac{FT} internals can improve model safety and efficiency, it could also enable adversarial exploitation. To mitigate this risk, we advocate for responsible disclosure practices and recommend that future work on efficient \ac{FT} methods include built-in safety mechanisms.

\section*{Acknowledgments}
This research was supported by the University of Liverpool and the Engineering and Physical Sciences Research Council (EPSRC). We thank the School of Computer Science and Informatics at the University of Liverpool for providing access to computational resources. The authors also acknowledge the anonymous reviewers for their constructive suggestions.

Danushka Bollegala holds concurrent appointments as a Professor at University of Liverpool and as an Amazon Scholar. This paper describes work performed at the University of Liverpool and is not associated with Amazon.

\bibliography{reference}

\appendix

\section{Description of Notations} \label{desc:notation}
In \autoref{fig:Intro}, each panel shows the top-12 highest-\(|s_e|\) edges of the EAP causal-attribution graph for one task. The visual encoding is the same across all panels.

  \paragraph{Nodes} are the transformer components touched by those edges, colour-coded by category:
  \swatch{CFE8B4}{input} (embedding),
  \swatch{EAD9BE}{m\(\ell\)} (MLP of layer~\(\ell\)),
  \swatch{C9D4EC}{a\(\ell\).h\(i\)} (attention head~\(i\) of layer~\(\ell\); Q/K/V ports collapsed), and
  \swatch{F1CFE0}{logits} (unembedding).

  \paragraph{Edges} are coloured by the sign of their attribution score \(s_e\): \textcolor[HTML]{C0392B}{red} for
  \(s_e>0\) (the edge helps the task loss) and \textcolor[HTML]{2E6F9E}{blue} for \(s_e<0\) (the edge hurts
  it). Line width is proportional to \(|s_e|\), so thicker edges are more causally important to the task; the thickness scale is shared across all panels, so importance is also comparable across tasks.
\begin{table*}[t]
\centering
\scriptsize 
\renewcommand{\arraystretch}{1.3} 

\begin{tabularx}{0.9\textwidth}{@{}p{0.18\textwidth}>{\raggedright\arraybackslash}X>{\raggedright\arraybackslash}X@{}}
\toprule
\textbf{Task Information} & \textbf{Original Input Sequence} & \textbf{Corrupted Input Sequence} \\ 

\midrule
\multicolumn{3}{c}{\cellcolor{gray!10}\textbf{\textit{Task 1: Sentiment Classification (SST-2/Yelp)}}} \\ 
\multicolumn{3}{l}{\textit{\textbf{Info Column:} Sentiment Label; \textbf{Input Column:} Review Text}} \\
\midrule
\textbf{Label: Negative} & a sometimes \textbf{tedious} film. & a sometimes \textbf{X} film. \\
\addlinespace
\textbf{Label: Positive} & among the year's most \textbf{intriguing} explorations of alienation. & among the year's most \textbf{X} explorations of alienation. \\
\midrule
\multicolumn{3}{c}{\cellcolor{gray!10}\textbf{\textit{Task 2: Machine Translation (Tatoeba/KDE4)}}} \\
\multicolumn{3}{l}{\textit{\textbf{Info Column:} Target Translation (French); \textbf{Input Column:} Source English}} \\
\midrule
\textbf{Target:} Il est resté silencieux. & He remained dumb. & x x xx \\
\midrule
\multicolumn{3}{c}{\cellcolor{gray!10}\textbf{\textit{Task 3: Question Answering (CoQA/SQuAD)}}} \\
\multicolumn{3}{l}{\textit{\textbf{Info Column:} Question; \textbf{Input Column:} Context Snippet (Answer Span Bolded)}} \\
\midrule
\textbf{Q:} What color was Cotton? & ... lived a little \textbf{white} kitten named Cotton. Cotton lived high up... & ... lived a little  \textbf{X} kitten named Cotton. Cotton lived high up... \\
\bottomrule
\end{tabularx}
\caption{Examples of corrupted samples across three downstream tasks. 
}
\label{tab:all_corrupted_examples}
\end{table*}

\section{Dataset Description} \label{ap:dataset}
\paragraph{Sentiment Classification Dataset.} In the sentiment classification task, we implemented two datasets:  Large Yelp Review (Yelp) Dataset \cite{yelp_polarity} and Stanford Sentiment Treebank (SST-2) \cite{sst2}. The Yelp dataset is used for binary sentiment classification; it contains $560,000$ training data and $38,000$ test data of highly polar Yelp reviews. The dataset is extracted from the Yelp Dataset Challenge by 2015 data. SST-2 corpus, which we use the version integrated into the GLUE benchmark \cite{glue}, consists of movie reviews originally extracted from the Rotten Tomatoes website. It provides a binary classification benchmark consisting of $67,349$ training, $872$ validation and $1,821$ test samples. In both datasets, each sentence is labelled as either positive or negative. For Yelp, we first select $11,000$ examples from the training split and apply a 90/10 split, resulting in $9,900$ examples used for fine-tuning. Evaluation is performed on $1,000$ examples selected from the test split. For SST-2, the GLUE test labels are not public, so we evaluate on the full labeled validation split. Training uses the first $15,000$ examples of the SST-2 train split. 

\paragraph{Question Answering Dataset.} In the question answering task, we used two benchmark reading-comprehension-style question answering datasets: the Stanford Question Answering Dataset (SQuAD) v1.1 \cite{squad} and CoQA \cite{coqa}. Both datasets provide a context passage as part of the input, and the answer is expected to be grounded in the given context rather than generated solely from the model’s parametric knowledge. In the train/validation splits used in our experiments, SQuAD v1.1 contains 87,599 training and $10,570$ validation question-answer pairs, totaling $98,169$ QAs. The dataset is based on Wikipedia articles, where each question is answered using a text span from the corresponding context passage. CoQA contains $108,647$ training and $7,983$ validation QAs, totaling $116,630$ QAs. It is designed for conversational question answering across multiple domains and emphasises multi-turn interactions with context-dependent responses. For SQuAD, we select $11,000$ examples from the training split and apply a 90/10 split, resulting in 9,900 examples used for fine-tuning. We evaluate on $1,000$ examples from the SQuAD validation split. For CoQA, we flatten the training examples, select $40,000$ instances, and apply a 90/10 split, resulting in $36,000$ examples used for fine-tuning. We evaluate on $1,000$ flattened examples from the CoQA validation split.

\paragraph{Machine Translation Dataset.}
We employed two parallel corpora for machine translation model: the KDE4 dataset \cite{tiedemann-2012-parallel} and Tatoeba dataset \cite{artetxe-schwenk-2019-massively}. The KDE4 dataset consists of a bilingual corpus derived from the localisation files of the KDE Software Environment, with English and French pairs of sentences. It has almost $200,000$ sentence pairs. The English sentences in the KDE4 dataset have an average length of approximately $12.4$ words, while the French translations average $14.7$ words, reflecting the slightly more verbose nature of French in technical
contexts. The Tatoeba dataset is a community-driven, open-source collection of sentence translations across multiple languages, including English and French. It is widely used to evaluate machine translation models because of its diversity and naturalistic content. This dataset contains approximately $300,000$ English-French sentence pairs. For KDE4, we select 30,000 examples and apply a 90/10 split, resulting in $27,000$ examples used for fine-tuning. Evaluation is conducted on a disjoint slice of $1,000$ examples from the KDE4 training corpus. For Tatoeba, we select $40,000$ examples and apply a 90/10 split, resulting in $36,000$ examples used for fine-tuning. Evaluation is conducted on a disjoint slice of $1,000$ examples from the Tatoeba training corpus.

\section{Corrupted Dataset Generation Details} \label{app:corrupted_data}

To analyse the localisation of task-specific knowledge, we constructed corrupted versions of the datasets by masking tokens that are critical for task performance. This section details the corruption strategy and provides qualitative examples for each task.

\paragraph{Sentiment Word Identification Details}
\label{app:sentiment_details}

Instead of using a fixed lexicon, we leveraged \texttt{Meta-Llama-3-8B-Instruct}\footnote{\url{https://huggingface.co/meta-llama/Meta-Llama-3-8B-Instruct}} to dynamically identify sentiment-bearing tokens relevant to the specific context. The process involved three steps:

\textbf{Sample Selection:} We first filtered for short sentences (5--10 tokens) from the validation set where our fine-tuned model originally made correct predictions. This ensures that the identified words are indeed responsible for the model's correct classification behaviour.

 \textbf{LLM Querying:} We constructed a few-shot prompt containing three examples (positive, negative, and neutral) to guide Llama-3 in extracting the most indicative sentiment words. The exact prompt template used is shown in \autoref{fig:prompt_template}.

 \textbf{Response Parsing:} The generated output was parsed to extract a comma-separated list of words. We strictly validated that the extracted words appeared in the original text and filtered out function words or tokens with fewer than 3 characters. Up to three unique words were selected per sample.

\begin{figure}[h]
\centering
\begin{promptbox}[title={Prompt Used for Sentiment Word Identification}]
\small
\textbf{System Instruction:} \\
You are a helpful assistant that extracts sentiment words from reviews.

\vspace{0.5em}
\hrule
\vspace{0.5em}

\textbf{User Input:} \\
Given that this review has a \placeholder{sentiment} sentiment, identify up to three words that most strongly indicate this \placeholder{sentiment} sentiment. Only select words that appear in the review text. Return the words in a comma-separated list. If no words strongly indicate the sentiment, return an empty list.

\textit{Examples:}
\begin{itemize}
    \item 
    \textbf{Review:} The food was amazing and the service was excellent! \\
    \textbf{Sentiment:} positive $\rightarrow$ \textbf{Output:} amazing, excellent
    \item \textbf{Review:} Terrible experience, never coming back. \\
    \textbf{Sentiment:} negative $\rightarrow$ \textbf{Output:} terrible
    \item \textbf{Review:} It was okay, nothing special. \\
    \textbf{Sentiment:} negative $\rightarrow$ \textbf{Output:} \textit{(empty)}
\end{itemize}

\vspace{0.5em}
\textit{Current Task:} \\
\textbf{Review:} \placeholder{input\_text} \\
\textbf{Sentiment:} \placeholder{label} \\
\textbf{Output:}
\end{promptbox}
\caption{The few-shot prompt template used to extract sentiment-sensitive words via Llama-3-8B-Instruct. We dynamically insert the specific sentiment label (positive/negative) and the review text into the placeholders (marked in blue) for each query.}
\label{fig:prompt_template}
\end{figure}

\autoref{tab:all_corrupted_examples} presents a comprehensive side-by-side comparison of original versus corrupted inputs across the three downstream tasks.

\begin{figure*}[htb]
    \centering
     \begin{subfigure}[b]{0.30\textwidth} 
\includegraphics[width=\linewidth, trim={6pt 6pt 6pt 6pt}, clip]{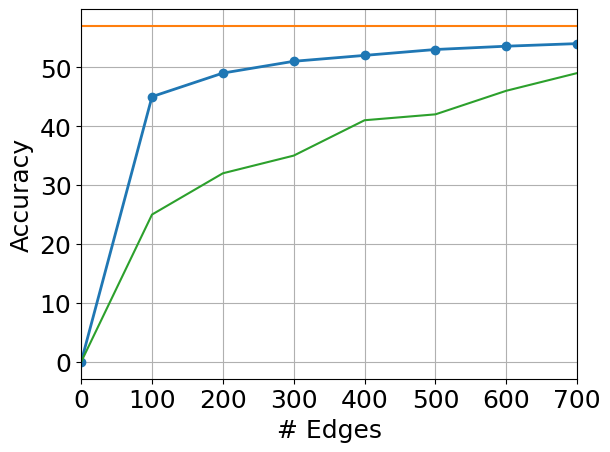}
\caption{GPT2 Small on Yelp}
\end{subfigure}
 \begin{subfigure}[b]{0.30\textwidth} 
\includegraphics[width=\linewidth, trim={6pt 6pt 6pt 6pt}, clip]{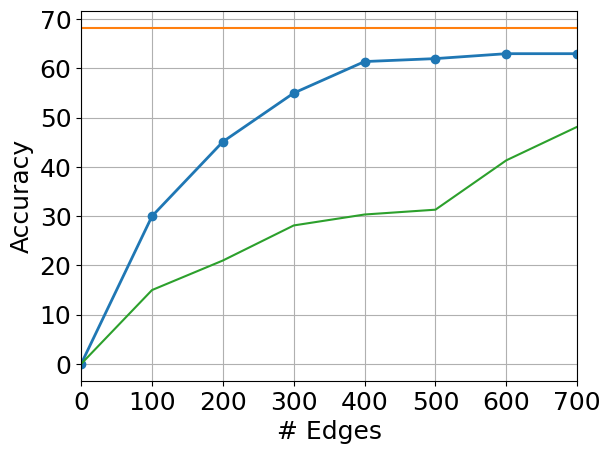}
\caption{Qwen2-0.5B on Yelp}
\end{subfigure}
 \begin{subfigure}[b]{0.30\textwidth} 
\includegraphics[width=\linewidth, trim={6pt 6pt 6pt 6pt}, clip]{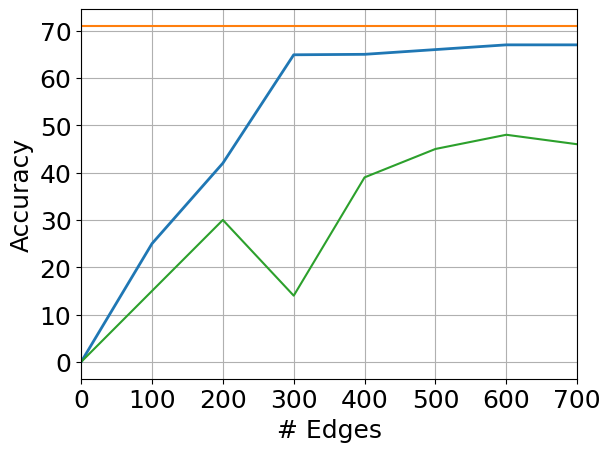}
\caption{Llama-3.2-1B on Yelp}
\end{subfigure}
 \begin{subfigure}[b]{0.30\textwidth} 
\includegraphics[width=\linewidth, trim={6pt 6pt 6pt 6pt}, clip]{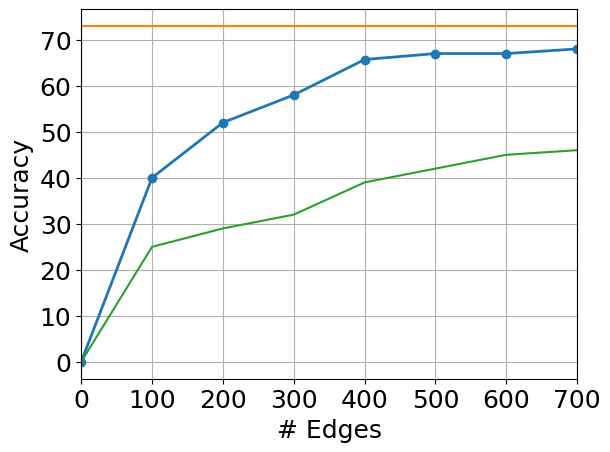}
\caption{Llama-2-7B on Yelp}
\end{subfigure}
 \begin{subfigure}[b]{0.30\textwidth} 
\includegraphics[width=\linewidth, trim={6pt 6pt 6pt 6pt}, clip]{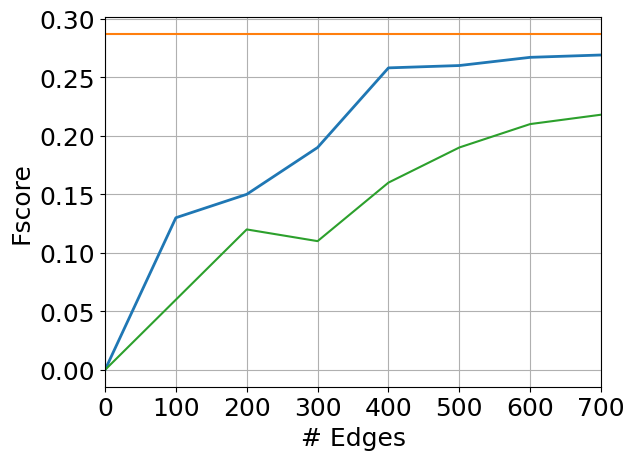}
\caption{GPT2 Small on SQuAD}
\end{subfigure}
\begin{subfigure}[b]{0.30\textwidth} 
\includegraphics[width=\linewidth, trim={6pt 6pt 6pt 6pt}, clip]{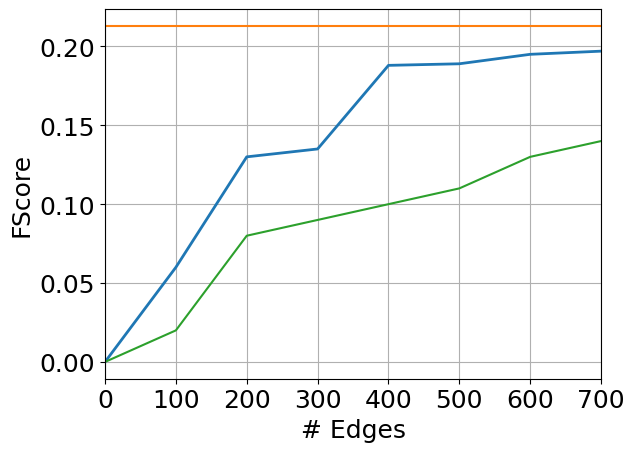}
\caption{Qwen2-0.5B on SQuAD}
\end{subfigure}
\begin{subfigure}[b]{0.30\textwidth} 
\includegraphics[width=\linewidth, trim={6pt 6pt 6pt 6pt}, clip]{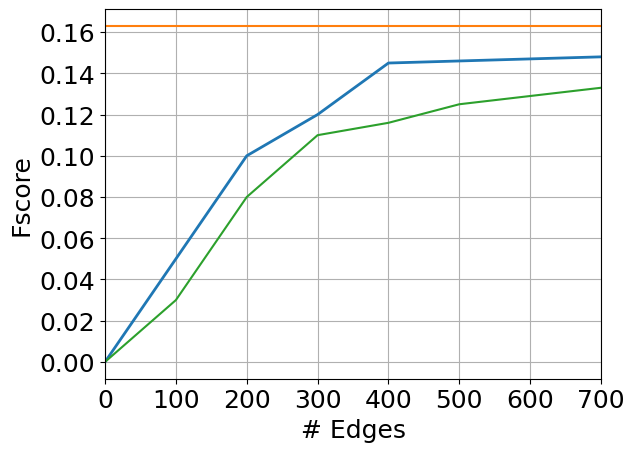}
\caption{Llama-3.2-1B on SQuAD}
\end{subfigure}
\begin{subfigure}[b]{0.30\textwidth} 
\includegraphics[width=\linewidth, trim={6pt 6pt 6pt 6pt}, clip]{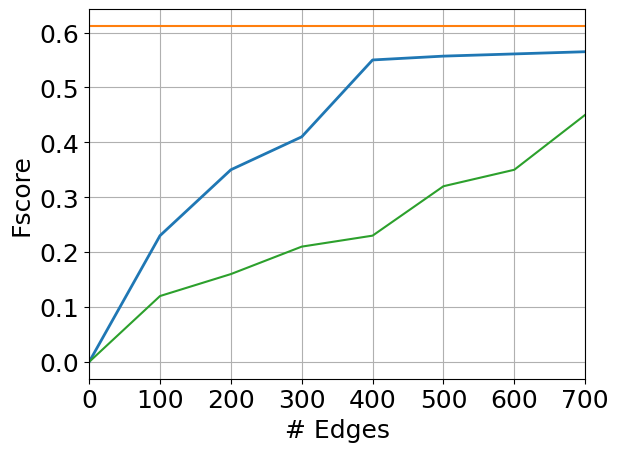}
\caption{Llama-2-7B on SQuAD}
\end{subfigure}
 \begin{subfigure}[b]{0.30\textwidth} 
\includegraphics[width=\linewidth, trim={6pt 6pt 6pt 6pt}, clip]{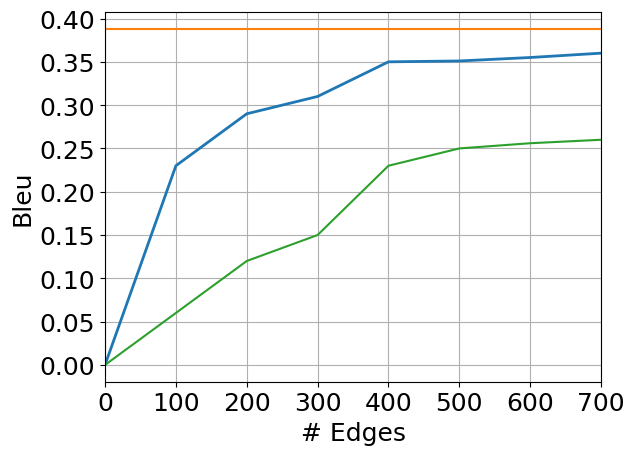}
\caption{GPT2 Small on KDE4}
\end{subfigure}
 \begin{subfigure}[b]{0.30\textwidth} 
\includegraphics[width=\linewidth, trim={6pt 6pt 6pt 6pt}, clip]{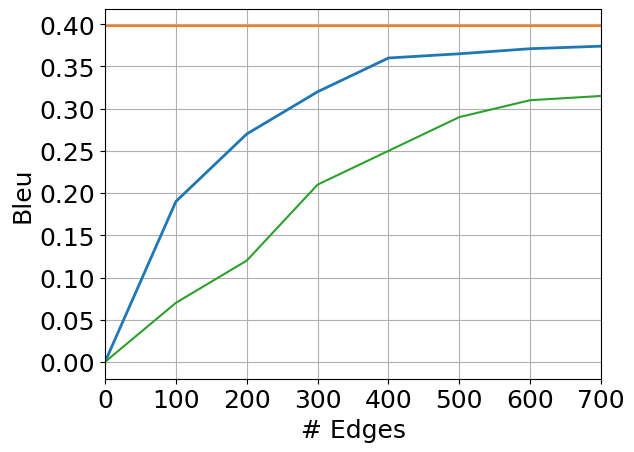}
\caption{Qwen2-0.5B on KDE4}
\end{subfigure}
 \begin{subfigure}[b]{0.30\textwidth} 
\includegraphics[width=\linewidth, trim={6pt 6pt 6pt 6pt}, clip]{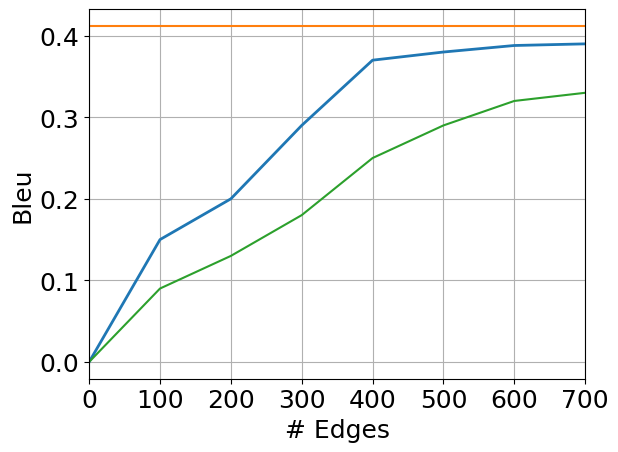}
\caption{Llama-3.2-1B on KDE4}
\end{subfigure}
 \begin{subfigure}[b]{0.30\textwidth} 
\includegraphics[width=\linewidth, trim={6pt 6pt 6pt 6pt}, clip]{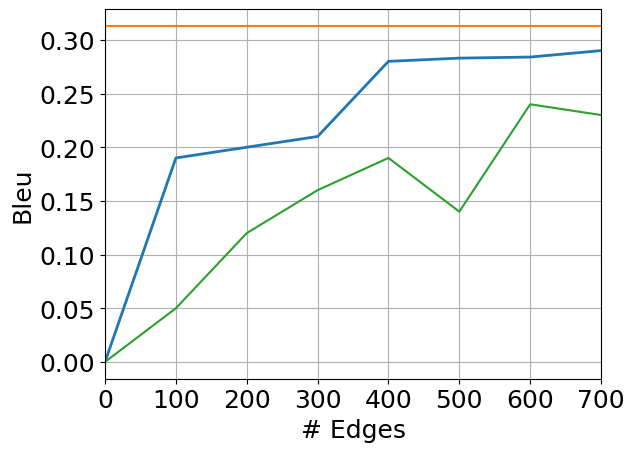}
\caption{Llama-2-7B on KDE4}
\end{subfigure}
    \caption{GPT-2 Small, Llama-3.2-1B, Qwen2-0.5B and Llama-2-7B performance with using only top-K edges obtained from EAP. The first row shows the performance on Yelp data, the second row shows the performance on SQuAD data, the third row shows the performance on KDE4 data. Orange line shows the performance of the finetuned model, blue line shows the performance with variation of top-k edges and green line shows the performance using randomly selected k edges.}
    \label{fig:ablation-top-edges}
\end{figure*}

\section{Validation of EAP-identified components} 
\label{ap:validation eap}

This section provides three additional analyses to demonstrate the faithfulness and robustness of the EAP edge rankings: a) performance comparison of retaining the top-\(k\) EAP-ranked edges, retaining the same number of randomly selected edges and retaining all the edges at different k values, which is to prove the rationale for setting the top-400 edges in our research; b) performance comparison that patching the activations of the top-400 EAP-ranked edges against that caused by patching on these same number of randomly selected components, which is to strengthen the core causal claim; c) compare the Spearman rank correlation of resulting edge-important scores between corruption token "<mask>" and other tokens, which is to evaluate the robustness of corruption token selection. 


\subsection{Sufficiency and the Choice of $k$}
\label{ap:ablation}
\autoref{fig:ablation-top-edges} shows the effect of using only top-k edges (k = {100, 200, 300, 400, 500, 600, 700}) on the performance of GPT-2 small and Llama-3.2-1B, Qwen2-0.5B and Llama-2-7B in Yelp dataset based sentiment classification task. More specifically, we retain only the top-k edges identified for each task and set the contributions of all remaining edges to zero. It can be observed that using only $400$ edges the $90\%$ of the model's original performance can be attained. This shows the faithfulness of the top-400 edges identified by EAP. We can also observe that the identified top-k edges have always performed better than the randomly selected k edges.

\subsection{Necessity of Corrupting the Top-400 Edges}
\label{ap:necessity}
To verify that the observed degradation is specific to the top-400 EAP-ranked edges, rather than a consequence of intervening on a large number of edges, in \autoref{tab:random-baseline}, we compare the performance drop caused by intervening on these edges against that caused by intervening on the same number of randomly selected edges. In both cases, we intervene by activation patching. For each selected edge, its clean-run contribution to the child input is replaced by the corresponding contribution from the corrupted run. The random baseline is constructed by uniformly sampling 400 edges from the full set of model edges, and the results are averaged over 20 random seeds.
\begin{table}[t]
\centering
\small
\begin{tabular}{llccc}
\toprule
\textbf{Model} & \textbf{Task} & \textbf{EAP} & \textbf{Random} & \textbf{$\Delta$} \\
\midrule
\multirow{3}{*}{GPT-2 Small}
  & Sent (Yelp)  & 85\% & 61\% & 24 \\
  & QA (SQuAD)   & 80\% & 55\% & 25 \\
  & MT (KDE4)    & 80\% & 67\% & 13 \\
\midrule
\multirow{3}{*}{Llama-2-7B}
  & Sent (Yelp)  & 81\% & 58\% & 23 \\
  & QA (SQuAD)   & 84\% & 52\% & 32 \\
  & MT (KDE4)    & 79\% & 54\% & 25 \\
\midrule
\multirow{3}{*}{Qwen2-0.5B}
  & Sent (Yelp)  & 83\% & 59\% & 24 \\
  & QA (SQuAD)   & 82\% & 55\% & 27 \\
  & MT (KDE4)    & 84\% & 56\% & 28 \\
\midrule
\multirow{3}{*}{Llama-3.2-1B}
  & Sent (Yelp)  & 81\% & 61\% & 20 \\
  & QA (SQuAD)   & 86\% & 52\% & 34 \\
  & MT (KDE4)    & 82\% & 44\% & 38 \\
\bottomrule
\end{tabular}
\caption{Performance drop caused by intervening on the top-400 EAP-ranked edges versus the same number of randomly selected edges. Random results are averaged over 20 seeds. $\Delta$ is the absolute difference in percentage points. The difference is significant for every model--task pair (two-sided Wilcoxon signed-rank test, $p < 0.05$).}
\label{tab:random-baseline}
\end{table}

\begin{table}[t]
\centering
\small
\begin{tabular}{lc}
\toprule
\textbf{Corruption token} & \textbf{Spearman $\rho$ vs.\ \texttt{<mask>}} \\
\midrule
\texttt{neutral} & 0.71 \\
\texttt{blank}   & 0.72 \\
\texttt{abc}     & 0.71 \\
\bottomrule
\end{tabular}
\caption{Spearman rank correlation between edge-importance scores obtained with alternative corruption tokens and those obtained with \texttt{<mask>}, computed over all edges of GPT-2 Small on the sentiment classification task.}
\label{tab:corruption-token}
\end{table}

\subsection{Robustness to the Corruption Token}
\label{app:corruption token}
In \autoref{tab:corruption-token}, we run \ac{EAP} on GPT-2 fine-tuned on the sentiment classification task while varying the corruption token (i.e., "neutral", "blank", "abc"), and measure the Spearman rank correlation between the resulting edge-importance scores over the full model. The edge rankings are highly correlated across tokens. It can be seen that for all the corrupted token variant it has a strong correlation with the edges obtained using "<mask>" token.

\section{Entropy of the Attention-Change and EAP Distributions}
\label{app:entropy}
We summarise how localised each distribution in \autoref{fig:eap_attention} is by a single entropy value, computed as follows. For a model with $L$ layers, each distribution provides one value per layer: $x_1,\dots,x_L$, where $x_\ell$ is either the average attention change $\Delta A_\ell$ (\autoref{eq:change}) or the summed \ac{EAP} score at layer $\ell$. We treat these per-layer values as a distribution over layers by normalising them,
\[
  p_\ell = \frac{x_\ell}{\sum_{j=1}^{L} x_j},
  \qquad \sum_{\ell=1}^{L} p_\ell = 1,
\]
and then collapse the $L$ per-layer values into a single scalar via the Shannon entropy (in bits) and its normalised form,
\[
  H = -\sum_{\ell=1}^{L} p_\ell \log_2 p_\ell,
  \qquad
  \tilde{H} = \frac{H}{\log_2 L},
\]
with the convention $0\log_2 0 = 0$. The summation over layers is what reduces each per-layer distribution to one number, so every distribution yields a single entropy value. Normalising by $\log_2 L$ (the maximum entropy, attained by a uniform distribution) maps $\tilde{H}$ to
$[0,1]$ and makes it comparable across models of different depths: $\tilde{H}\to 1$ means the change is spread evenly across all layers, while $\tilde{H}\to 0$ means it is concentrated in a few layers. Thus, each subplot is summarised by the pair $(\tilde{H}_{\text{attn}},\,\tilde{H}_{\text{EAP}})$ reported in the caption of \autoref{fig:eap_attention}. Across all models and tasks, $\tilde{H}_{\text{EAP}}$ is substantially lower than $\tilde{H}_{\text{attn}}$, confirming that \ac{EAP}-identified components are concentrated in a small number of layers.

\section{Hyperparameter Setup in \ac{FT}} \label{ap:hyperparameter}
To verify our findings across various settings, we conduct experiments on four language model families: \textbf{GPT-2 Small (124M)}, \textbf{Qwen-2-0.5B}, \textbf{Llama-3.2-1B} and \textbf{Llama-2-7B}. All experiments were run on a single NVIDIA A100 80GB GPU. The total computational budget was approximately 14 GPU hours. The detailed fine-tuning parameters are in \autoref{tab:all_params}.

\begin{table}[htb]
  \centering
  \small
  \resizebox{\columnwidth}{!}{
  \begin{tabular}{lcccc}
  \toprule
  \textbf{Parameter} & \textbf{Llama-2 (7B)} & \textbf{Llama-3.2 (1B)} & \textbf{Qwen2 (0.5B)} & \textbf{GPT-2 (Small)} \\
  \midrule
  \textit{Core Optimization} \\
  Optimizer         & AdamW (fused) & AdamW (fused) & AdamW (fused) & AdamW (fused) \\
  Learning Rate     & $2\times10^{-5}$ & $2\times10^{-5}$ & $2\times10^{-5}$ & $5\times10^{-5}$\textsuperscript{$\dagger$} \\
  Weight Decay      & 0.0 & 0.0 & 0.0 & 0.0 \\
  Warmup Ratio      & 0.03 & 0.03 & 0.03 & 0.03 \\
  LR Scheduler      & Cosine & Cosine & Cosine & Cosine \\
  Precision         & bf16 & bf16 & bf16 & bf16 \\
  \midrule
  \textit{Effective Batch Size} \\
  SST-2 / Yelp      & 16 / 8 & 32 & 32 / 16 & 32 \\
  QA (SQuAD/CoQA)   & 8 & 32 & 32 / 16 & 32 \\
  MT (KDE4/Tatoeba) & 8 & 32 & 32 & 32 \\
  \midrule
  \textit{Training Epochs} \\
  SST-2 / Yelp      & 2 & 3 & 3 & 3 \\
  QA (SQuAD/CoQA)   & 2 / 1 & 2 & 2 & 3 \\
  MT (KDE4/Tatoeba) & 1 & 3 & 3 & 3 \\
  \midrule
  \textit{Sequence Length} \\
  SST-2 / Yelp      & 256 & 512 & 512 & 512 \\
  QA (SQuAD/CoQA)   & 1024 & 1024 & 1024 & 1024 \\
  MT (KDE4/Tatoeba) & 256 & 256 & 256 & 256 \\
  \bottomrule
  \end{tabular}}
  \caption{Complete hyperparameter settings across models and tasks. All four models are fully fine-tuned. Effective batch size is per-device batch size $\times$ gradient-accumulation steps. \textsuperscript{$\dagger$}GPT-2
  SQuAD/CoQA used $2\times10^{-5}$.}
  \label{tab:all_params}
  \end{table}

     \begin{table*}[htb]
   \centering
  \footnotesize
  \resizebox{\textwidth}{!}{
  \begin{tabular}{|l|c|c|c|c|c|c|c|c|c|}
  \hline
  \multicolumn{10}{|c|}{Similar Task Analysis}\\
  \hline
\multicolumn{2}{|c|}{\textbf{Task Pair}}&\multicolumn{2}{|c|}{\textbf{GPT2}} &  \multicolumn{2}{|c|}{\textbf{Llama3.21B}}  &  \multicolumn{2}{|c|}{\textbf{Qwen2.05B}} &  \multicolumn{2}{|c|}{\textbf{Llama-2-7B}}\\
   \hline
 Fine-tuning Task & Evaluation Task & Perf $\Delta$ & Overlap  $\Delta$& Perf $\Delta$ & Overlap $\Delta$ & Perf $\Delta$ & Overlap  $\Delta$& Perf $\Delta$ & Overlap $\Delta$\\
   \hline
  Yelp & Yelp&34\% $\uparrow$& 100 \%& 33.81\% $\uparrow$& 100\%& 25.1\% $\uparrow$& 100\% & 91\% $\uparrow$ & 100\%\\
 Yelp & SST-2& 47\% $\uparrow$&45.25\%& 52.2\% $\uparrow$&39.75\% & 23.51\% $\uparrow$& 33.50\%& 94.5\% $\uparrow$ & 16.75\%\\
 SST-2 & SST-2&52.81\%$\uparrow$&100\%& 21.92\% $\uparrow$&100\%& 31.77\% $\uparrow$& 100\% &96.44\%$\uparrow$& 100\%\\
 SST-2 & Yelp&29.39\% $\uparrow$&37.25\% &8.41\% $\uparrow$& 40.5\% & 16.7\% $\uparrow$& 35.00 \% & 89.2\% $\uparrow$ & 8.00 \%\\
  SQuAD & SQuAD&535\% $\uparrow$&100\%&417.54\% $\uparrow$& 100\%& 65.55\% $\uparrow$& 100\% & 81.23\%  $\uparrow$& 100\%\\
 SQuAD &CoQA&216\% $\uparrow$ &37.00\%&197.71\% $\uparrow$&30.75\%& 41.16\% $\uparrow$& 35.50\% & 27.84\%  $\uparrow$& 45.50\%\\
CoQA & CoQA & 287\% $\uparrow$ &100\%&203\% $\uparrow$&100\%& 106.59 \% $\uparrow$& 100\% & 5.95\%$\uparrow$ & 100\%\\
 CoQA &  SQuAD &319\% $\uparrow$ & 36.50\% &312\% $\uparrow$ & 31.75\% & 56.99\% $\uparrow$ & 40.00 \% & 64.33\% $\uparrow$ & 51.25\%\\
 KDE4 & KDE4 & 441\% $\uparrow$& 100\%&188\% $\uparrow$&100\%&119.90\% $\uparrow$& 100\%  & 10.31\% $\uparrow$ & 100\%\\
  KDE4 &  Tatoeba & 206\% $\uparrow$& 47\%& 4.52\% $\uparrow$&30.25\%&4.13\% $\uparrow$&37.50\%&11.35\% $\uparrow$ & 31.75\% \\
  Tatoeba & Tatoeba& 384\% $\uparrow$&100\%&39.82\%&100\%&69.48\% $\uparrow$& 100\% & 24.53\% $\uparrow$ & 100\% \\
  Tatoeba & KDE4 & 13.83\% $\uparrow$& 47.00\%&76.84\% $\uparrow$&32.50\%&9.92\% $\downarrow$& 40.00\% & 9.33\% $\uparrow$ & 29.25\% \\
  \hline
  \hline
 \multicolumn{10}{|c|}{Dissimilar Task Analysis}\\
 \hline
Yelp & SQuAD&67\%& 32.50\%&81.71\% &26.75\% &9.50\% &30.50\% & 0.59\% & 44.50\% \\
Yelp & CoQA&9.64\%& 30.50\% & 52.81\% & 30.50\% & 38.18\%& 29.75\% & 2.8\%  & 39.75\% \\
Yelp & KDE4&54.61\%& 41.00\% &82.92\% & 26.75\% & 38.97\% & 31.00\%& 8.20\% & 27.75\% \\
Yelp & Tatoeba&22.63\%& 42.75\%& 87.3\% & 27\% &26.85\% & 31.25\% & 0.08\% & 14.75\%\\ 
SST-2& SQuAD & 47\% & 36\% &90.31\% & 32.25\%& 72.29\% & 36.25\% & 24\% & 7\% \\
SST-2& CoQA&48.2\%& 36.75\% & 61.33\% & 26\% & 64.62\%& 36.25\%& 6.7\% & 14.75\%\\
SST-2 & KDE4&71.21\%& 42.00\%& 91.54\%& 29.75\% & 88.22\%& 39.25\%&  23.76\%& 19.75\% \\
SST-2 & Tatoeba&33.15\%&41.50\% & 99.00\%& 34.25\% & 99.31\%& 39.75\%& 33.15\% & 19.75\% \\
SQuAD& Yelp& 83.91\%& 33.50\% & 8.41\% & 28.75\% &58.31\% &33.75\%  &2.2\%& 46.75\% \\
SQuAD& SST-2&4.26\% &31.75\% &10.31\% & 31.50\% & 44.38\% &35.50\%  &6.77\% & 6.25\% \\
SQuAD& KDE4& 24.63\% & 35.75\% & 6.91\% & 30.50\% & 41.82\%& 30.50\%  & 66.52\% & 29.00 \%\\
SQuAD& Tatoeba&21.35\% & 35.00\% &3.81\% & 28.25\% &37.07\% & 31.50\%  &87.81\%  & 3.50\%\\

CoQA& Yelp& 0.31\%& 33.25\% & 1.81\% & 26.25\% & 4.21\% & 29.02\%  & 4.51\% & 38.52\%\\
CoQA& SST-2& 3.63\%& 34.00\%& 4.67\%& 25.00\% & 3.78\% & 32.25\% & 1.03\% & 12.50\%\\
CoQA& KDE4& 35.83\% & 31.50\% & 12.63\%& 26.00\% & 31.46\% & 27.75\%  & 25.86\% & 24.75\%\\
CoQA& Tatoeba&21.33\% & 31.75\% & 7.91\% & 19.75\% &35.75\% & 27.50\%  &21.44\% & 5.25\% \\

KDE4& Yelp& 3.31\%& 39.50\% & 5.82\% & 26.25\% & 68.12 \% & 33.00\%  & 6.51\% & 30.25\% \\
KDE4& SST-2& 6.91\%& 41.75\% & 13.81\%& 34.50\% & 60.09\% & 41.50\% & 0.11 \% & 25.75\%\\
KDE4& CoQA& 33.72\% & 30.75\% & 16.01\%& 30.25\% & 4.38\%& 34.00\%  & 4.91\% & 28.25\% \\
KDE4& SQuAD& 42.61\%& 34.00\% &13.14\% & 31.25\% & 13.19\% & 35.50\%  & 8.86\% & 28.50\%\\

Tatoeba& Yelp&3.00\% & 40.50\%&  9.64\%& 26.25\%& 71.53\% & 35.25\% & 6.92\% & 23.00\% \\
Tatoeba& SST-2& 5.41\%& 40.25\% & 29.48\% & 31.50\% & 60.78\% & 41.75\%  & 0.00\% & 14.00\% \\
Tatoeba& CoQA& 54.33\%& 31.00\% &6.22\% &25.50\% & 58.47\%& 32.00\%  & 1.97\% & 14.75\% \\
Tatoeba& SQuAD& 25.61\%& 34.75\% &26.24\% & 28.00\% & 91.00\% & 32.50\%  &6.56\% & 5.00\% \\  \hline\end{tabular}}\caption{Cross-task Performance and Important Component Overlap Analysis for GPT-2 Small, Llama-3.2-1B, Qwen2-0.5B and Llama-2-7B.}\label{tab:performance_overlap}
  \end{table*}

\section{Component Distribution in Top-K Edges} \label{ap:fig:piechart}
\begin{figure*}[t]
     \centering
 \includegraphics[width=0.9\textwidth]{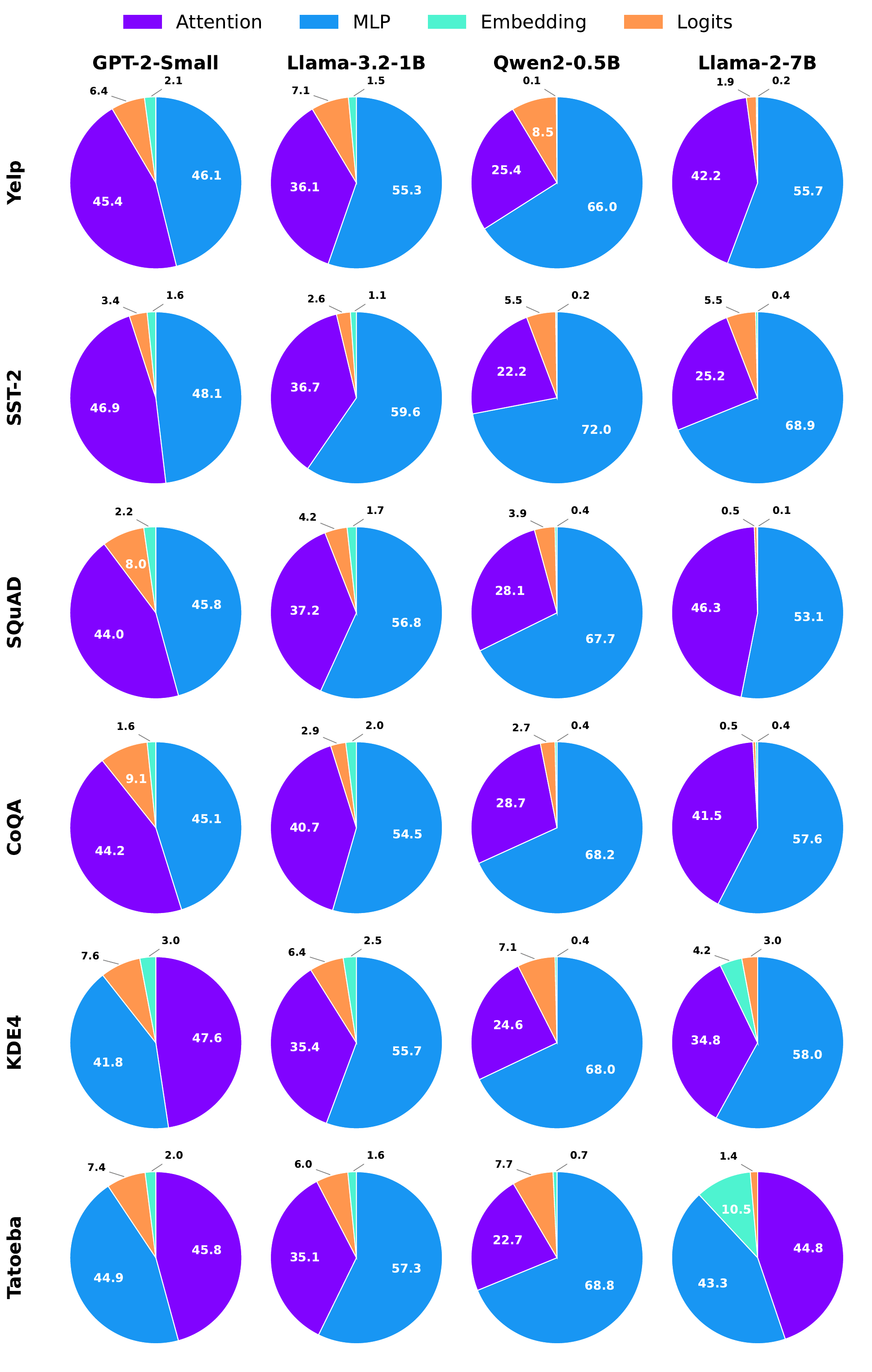}
     \caption{ Important component distribution percentage over top-400 \ac{EAP} edges, broken down by model(columns) and dataset(rows).The Llama-3.2-1B column on SST-2 / SQuAD / KDE4 reproduces the main-text \autoref{fig:component distribution}.  }
     \label{fig:all component distribution}
 \end{figure*}

\autoref{fig:all component distribution} shows component-wise distribution of top-400 edges across four models and six datasets.

\begin{figure*}[t]
     \centering
 \begin{subfigure}[b]{0.70\textwidth}       \includegraphics[width=\linewidth]{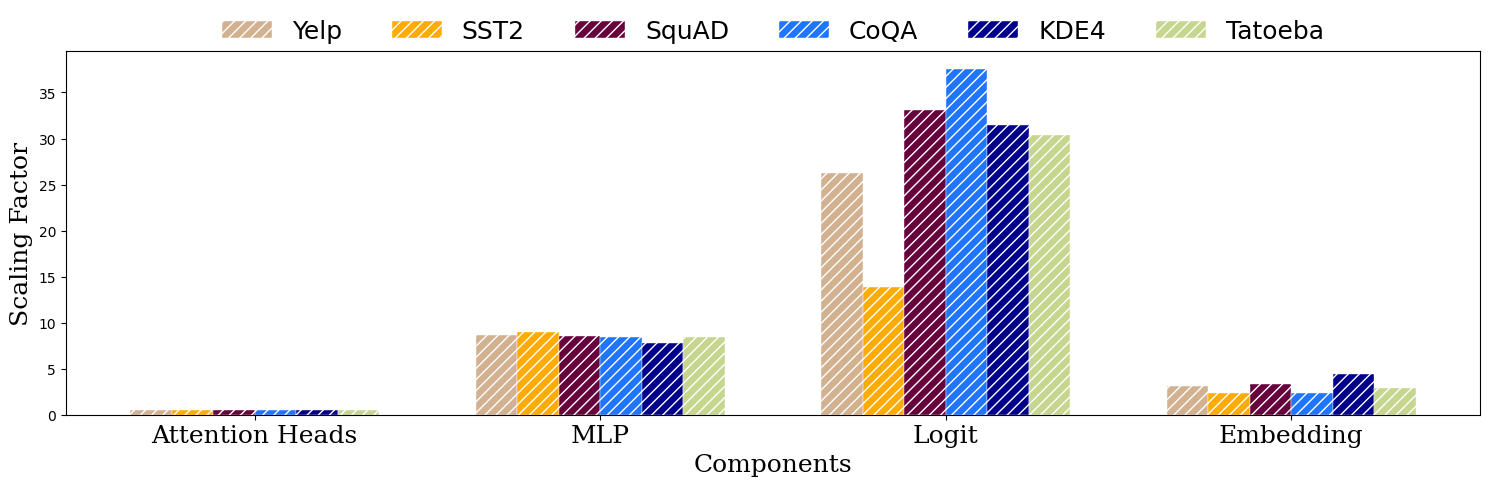}
 \caption{GPT2 Small}
 \end{subfigure}
  \begin{subfigure}[b]{0.70\textwidth}       \includegraphics[width=\linewidth]{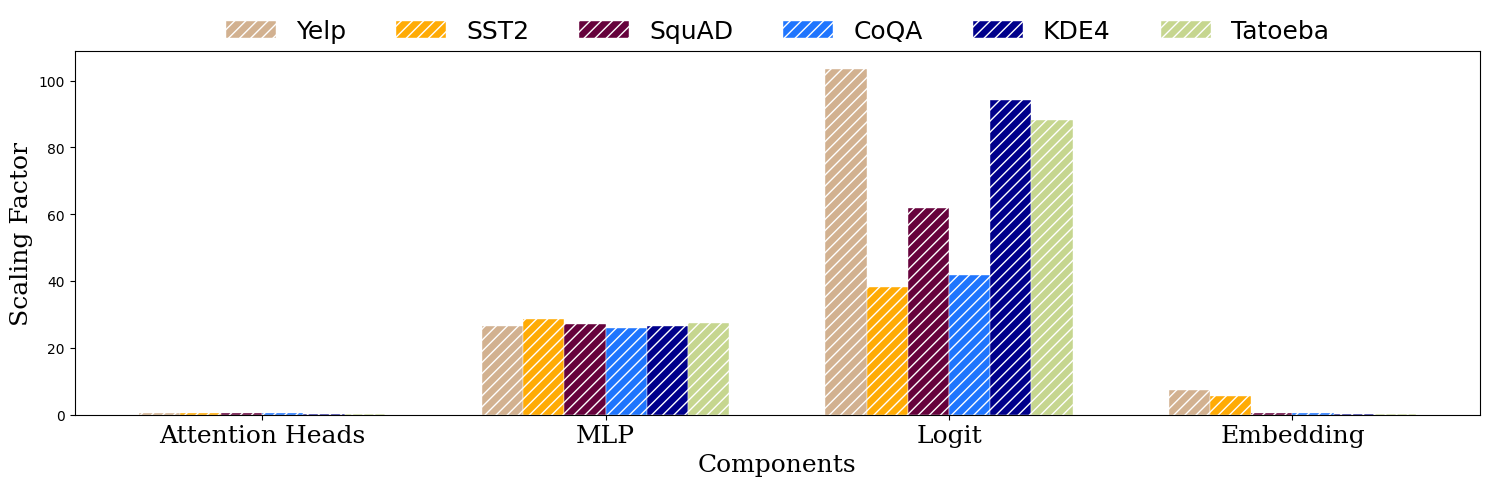}
  \caption{Llama-3.2-1B}
 \end{subfigure}
  \begin{subfigure}[b]{0.70\textwidth}       \includegraphics[width=\linewidth]{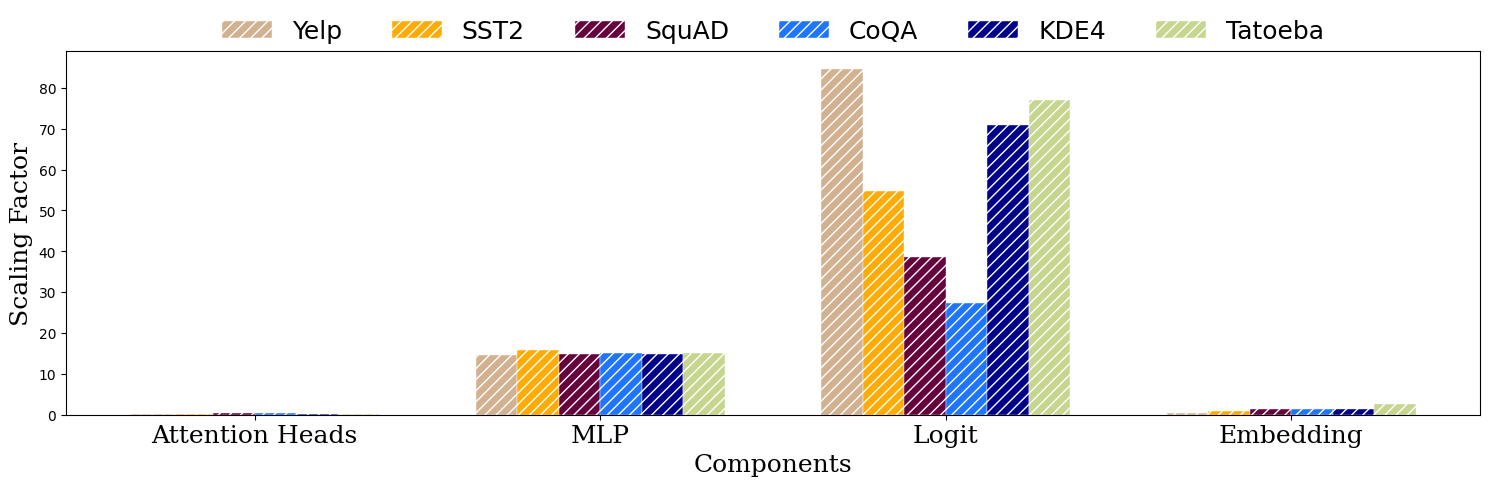}
  \caption{Qwen2-0.5B}
 \end{subfigure}
    \begin{subfigure}[b]{0.70\textwidth}       \includegraphics[width=\linewidth]{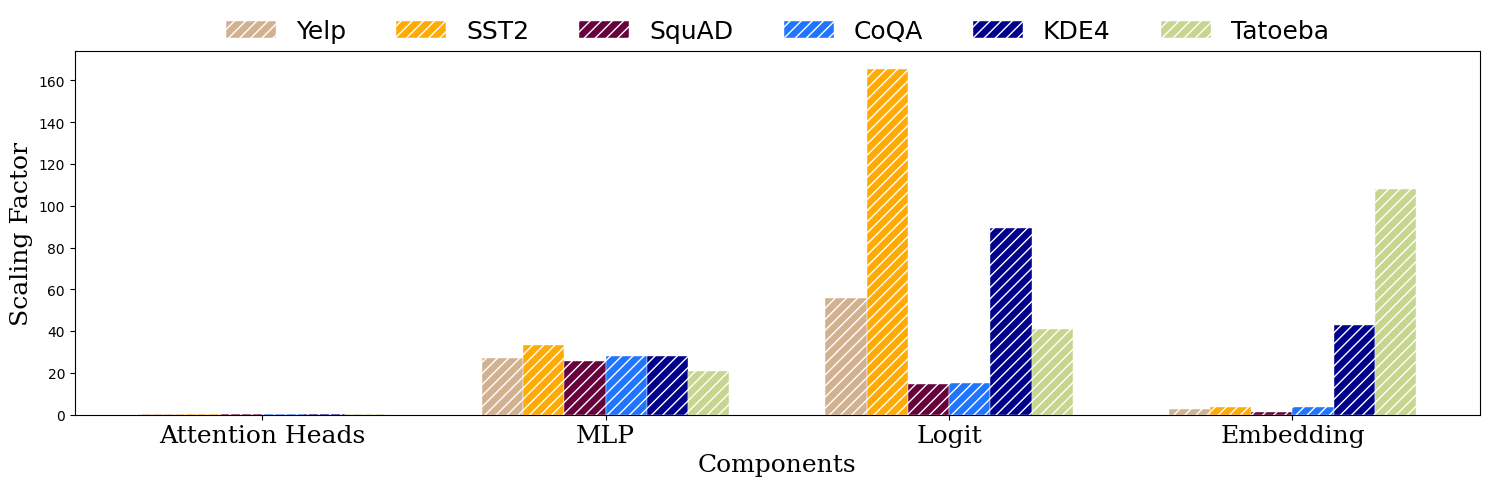}
    \caption{Llama-2-7B}
 \end{subfigure}  \caption{Architecture-normalised component composition of the top-400 EAP edges. For each model (a-d), bars show the scaling factor $r_c = f_c^{\text{top}}/f_c^{\text{model}}$, i.e.,\ the fraction of top-400 edges of component type $c$ divided by the fraction of type $c$ in the full model. Components are grouped along the $x$-axis (attention heads, MLP, logit, embedding) and the six bars per group correspond to the six tasks. A value above one indicates that the component type is selected more often than its baseline abundance would predict. See \autoref{norm:freq} for the definition. }

     \label{fig:norm component distribution}
 \end{figure*}

\paragraph{Normalised Frequency} \label{norm:freq}
\label{norm:freq}
The raw composition in \autoref{fig:all component distribution} is dominated by the component types that are simply most numerous in the model (e.g., MLP and attention nodes), which makes it hard to see whether a component type is selected more often than its baseline abundance would predict. To remove this frequency bias, we report an
architecture-normalised composition. Let $f_c^{\text{top}}$ be the fraction of the top-400 \ac{EAP} edges whose endpoint is of component type $c\in\{\text{attention, MLP, embedding, logit}\}$, and let $f_c^{\text{model}}$ be the fraction of all components of type $c$ in the full model. We define the \emph{scaling factor}
\[r_c = \frac{f_c^{\text{top}}}{f_c^{\text{model}}}, \]
which measures how strongly type $c$ is over- or under-represented among the \ac{EAP}-identified components relative to its natural abundance: $r_c>1$ means type $c$ is selected more often than chance, and $r_c<1$ means it is under-selected. \autoref{fig:norm component distribution} plots $r_c$ directly for each component type (grouped along the $x$-axis) and task (bars). Under this normalisation, the contributions of MLP and especially logits are amplified, indicating that although these types are not the most numerous overall, they are disproportionately selected as causally important.
\section{Cross Task Performance Transferability}
\label{ap:crosstask}
\begin{figure}[H]
    \centering
    \includegraphics[width=0.8\linewidth,trim={6pt 6pt 6pt 6pt}, clip]{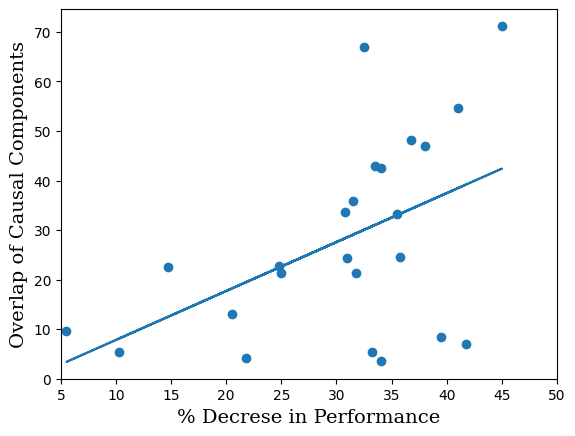}
    \caption{GPT2, $\rho=0.5$}
    \label{fig:gpt2-correlation}
\end{figure}
\section{Induction Head Detection Details}
\label{app:induction}
We detect induction heads with the synthetic repeated-sequence protocol of \citet{olsson2022}. We sample a random token sequence $X=[x_1,\dots,x_T]$ of length $T=50$, with token ids drawn uniformly from $[100,\,|V|)$ to avoid special and other reserved low-id tokens, and feed its twofold repetition $[X,X]$ of total length $2T=100$ to a \texttt{HookedTransformer} \citep{nanda2022transformerlens}, caching every layer's attention pattern. For each head we compute an \emph{induction score}: the mean attention, averaged over query positions $i\in\{T,\dots,2T{-}2\}$ in the repeated half, that the head places on key position $i-T+1$ ---the token immediately following the previous occurrence of the current token. A head is labelled an induction head by a top-$K$ rule on the sorted induction scores. Let $K_{\mathrm{elbow}}$ be the elbow of the sorted-score curve (the rank of maximum distance to the chord joining its first and last points), searched within the top $20\%$ of ranks; let $K_{0.1}$ and $K_{0.3}$ be the numbers of heads scoring above $0.1$ and $0.3$, respectively. We set
  \[
    K=\max\!\bigl(\min(K_{\mathrm{elbow}},\,K_{0.1}),\;K_{0.3}\bigr),
  \]
i.e.,\ the elbow cut, tightened by a minimum-score threshold of $0.1$, but never dropping a head whose score exceeds $0.3$ (a safety floor). The $K$ highest-scoring heads are retained. The same protocol is applied to the pre-trained and the fine-tuned model.

\section{Validation of the correlation claim}
\label{app:validation-correlation}
 
\autoref{sec:generalisability} reports a positive correlation between the overlap in \ac{EAP}-identified components and the degradation observed on Task 2 when the model is fine-tuned on Task 1. Because this evidence is correlational, it does not by itself establish that the shared components are responsible for the degradation. To test this mechanism directly, we perform an intervention: we identify the components that are shared between the two tasks, freeze them, and fine-tune only the remaining parameters on Task 1. If the overlapping components are indeed what drives the interference, protecting them during fine-tuning should recover part of the lost performance on Task 2.
 
\autoref{tab:freezing} reports the change in Task 2 performance under this modified fine-tuning procedure, relative to standard full fine-tuning on Task 1. Across all model and task-pair combinations, freezing the overlapping components improves Task~2 performance, with gains ranging from $2.05\%$ to $84.34\%$. 
 
\begin{table}[h]
\centering
\small
\begin{tabular}{llll}
\toprule
\textbf{Model} & \textbf{FT task} & \textbf{Test task} & \textbf{Change} \\
\midrule
\multirow{4}{*}{Llama-3.2-1B}
  & MT   & QA   & $\uparrow$ 3.54\%  \\
  & QA   & MT   & $\uparrow$ 45.74\% \\
  & Sent & MT   & $\uparrow$ 7.91\%  \\
  & Sent & QA   & $\uparrow$ 41.93\% \\
\midrule
\multirow{5}{*}{Qwen2-0.5B}
  & QA   & MT   & $\uparrow$ 84.34\% \\
  & MT   & QA   & $\uparrow$ 44.47\% \\
  & MT   & Sent & $\uparrow$ 2.05\%  \\
  & Sent & QA   & $\uparrow$ 83.82\% \\
  & Sent & MT   & $\uparrow$ 77.99\% \\
\bottomrule
\end{tabular}
\caption{Effect of freezing the overlapping EAP-identified components during
fine-tuning. The model is fine-tuned on Task~1 with the components shared
between Task~1 and Task~2 held fixed, and then evaluated on Task~2. The last
column reports the relative change in Task~2 performance compared to standard
full fine-tuning on Task~1. Performance on Task~2 improves in every
configuration.}
\label{tab:freezing}
\end{table}

\section{Layer-wise Centred Kernel Alignment Similarity}
\label{app:cka}
The probing analysis in \autoref{sec:finetuningdynamics} measures representational change by mapping intermediate hidden states through the model's output head (\autoref{eq:mrr}). This quantity is therefore output-mediated: it captures how recoverable the task output is from a given layer, rather than how much the geometry of the representation itself has changed. Since the EAP objective is also defined in terms of logit differences, we complement the probing analysis with a measure that does not pass through the output head.
 
\begin{figure*}[!h]
\centering
\includegraphics[width=\textwidth]{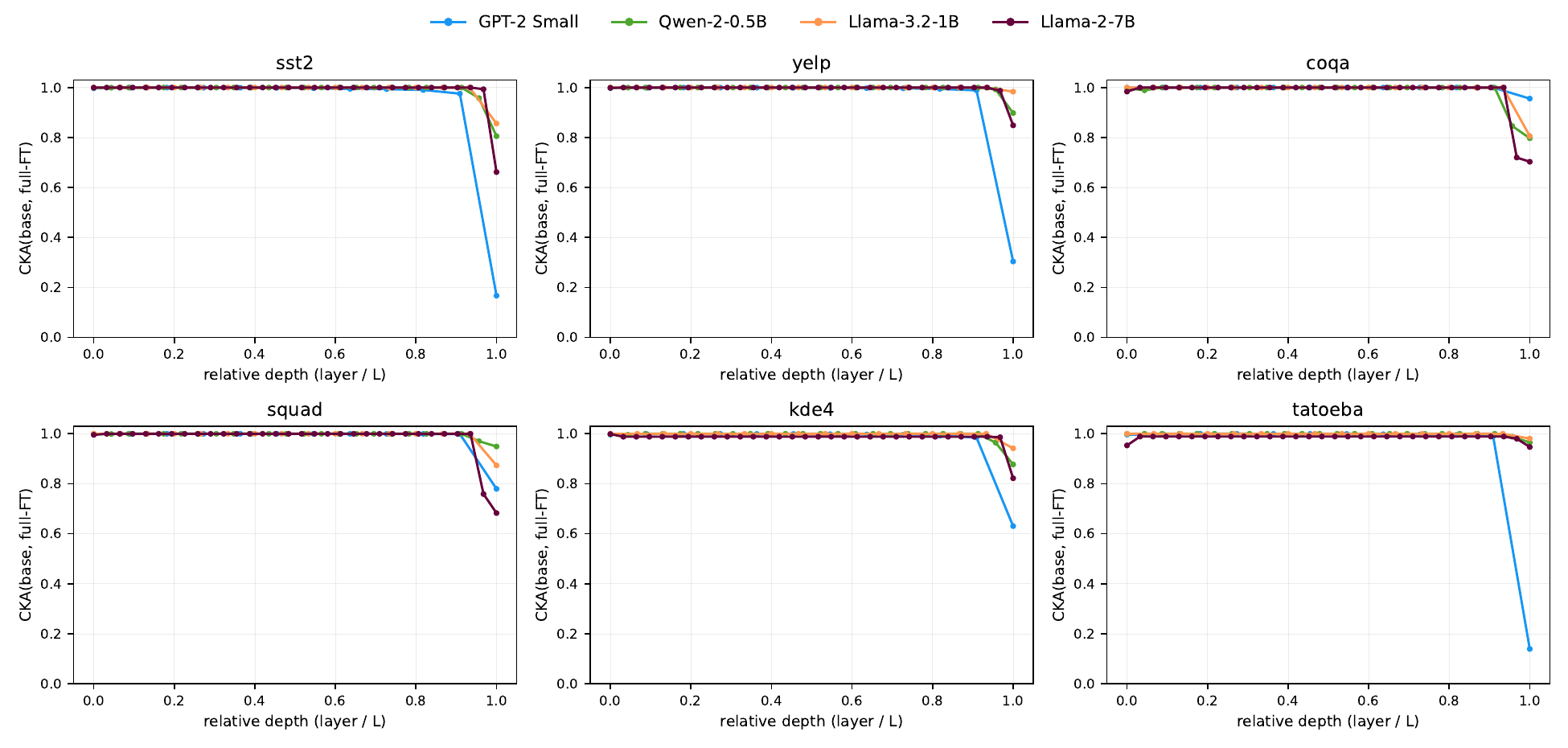}
\caption{Token-level CKA between the pre-trained and the fine-tuned model at each layer, for the four models (lines) across the six datasets (panels). The $x$-axis is relative depth (layer index divided by the number of layers), which allows models of different depths to be compared on a common scale. A CKA of $1$ means the representation at that layer is unchanged by fine-tuning; lower values indicate larger representational change. }
\label{fig:cka}
\end{figure*}

Specifically, we compute token-level hidden states Centred Kernel Alignment (CKA), a standard measure of similarity between layer-wise representations, between the pre-trained and fine-tuned models at every layer. A value of $1$ indicates that the representation at that layer is unchanged by fine-tuning, so lower values correspond to larger representational change. Because the four models differ in depth, we plot CKA against relative depth (layer index divided by the total number of layers), which makes the curves directly comparable across models. \autoref{fig:cka} shows the resulting curves for all models and datasets. Across all models and tasks, the most substantial layer-wise representational changes occur in the later layers (in the order of 0.7 to 0.9), specifically from approximately the 90th percentile layer to the final layer.

\section{More main results}
\label{app:rest-results}
\begin{figure*}[t]
  \centering  \begin{subfigure}[b]{0.325\textwidth}       \includegraphics[width=\linewidth]{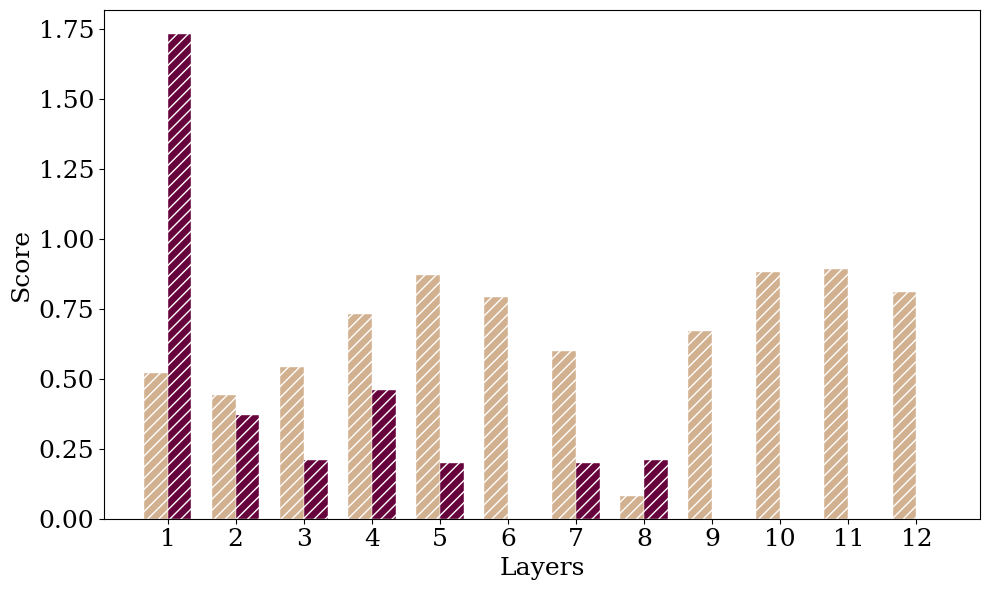}
    \caption{GPT2 on SST-2 ($0.97$,$0.62$)}     \end{subfigure}
    \begin{subfigure}[b]{0.325\textwidth}     \includegraphics[width=\linewidth]{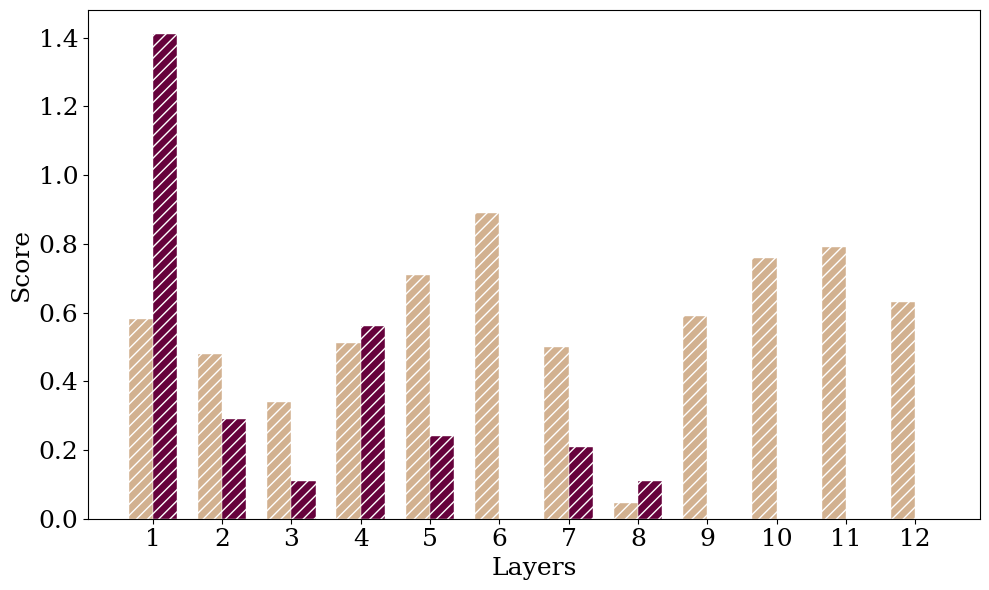}
    \caption{GPT2 on SQuAD ($0.96$,$0.62$)}     \end{subfigure}
\begin{subfigure}[b]{0.325\textwidth}
\includegraphics[width=\linewidth]{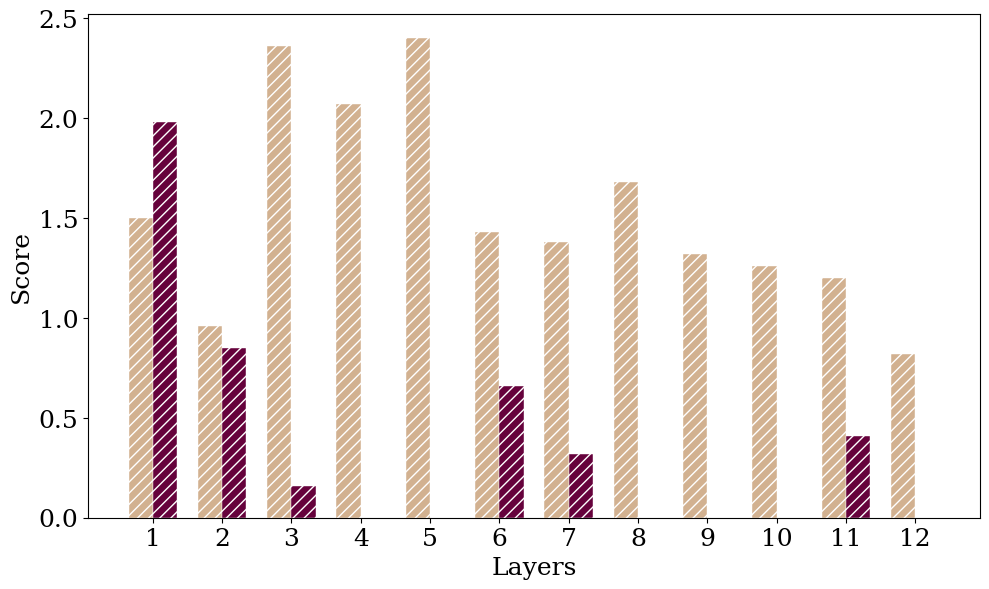}
        \caption{GPT2 on KDE4 ($0.98$,$0.60$)}
    \end{subfigure} 
\begin{subfigure}[b]{0.325\textwidth}
\includegraphics[width=\linewidth]{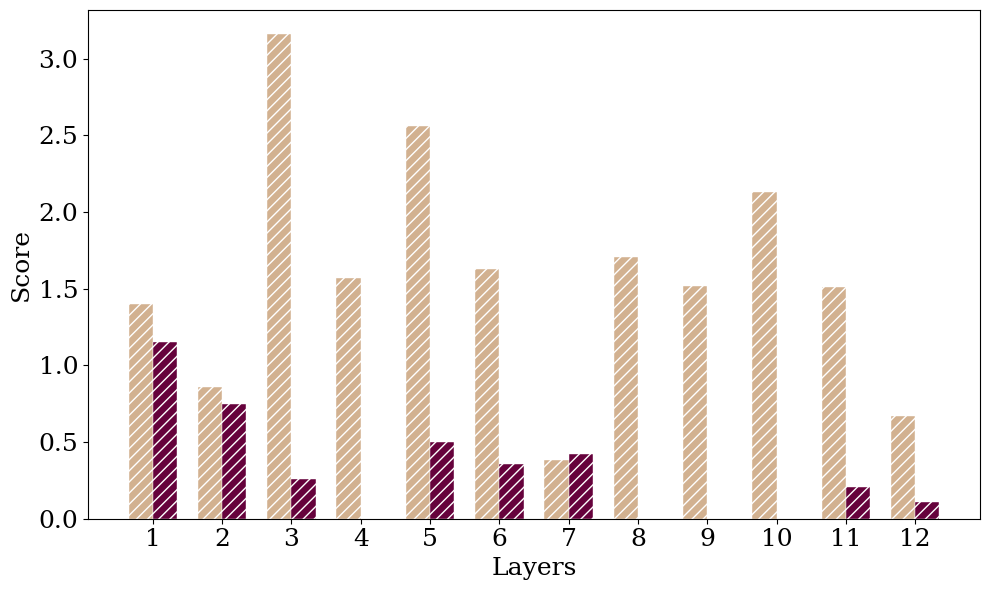}
    \caption{GPT2 on Yelp ($0.95$,$0.75$)}  
    \end{subfigure}
\begin{subfigure}[b]{0.325\textwidth}
        \includegraphics[width=\linewidth]{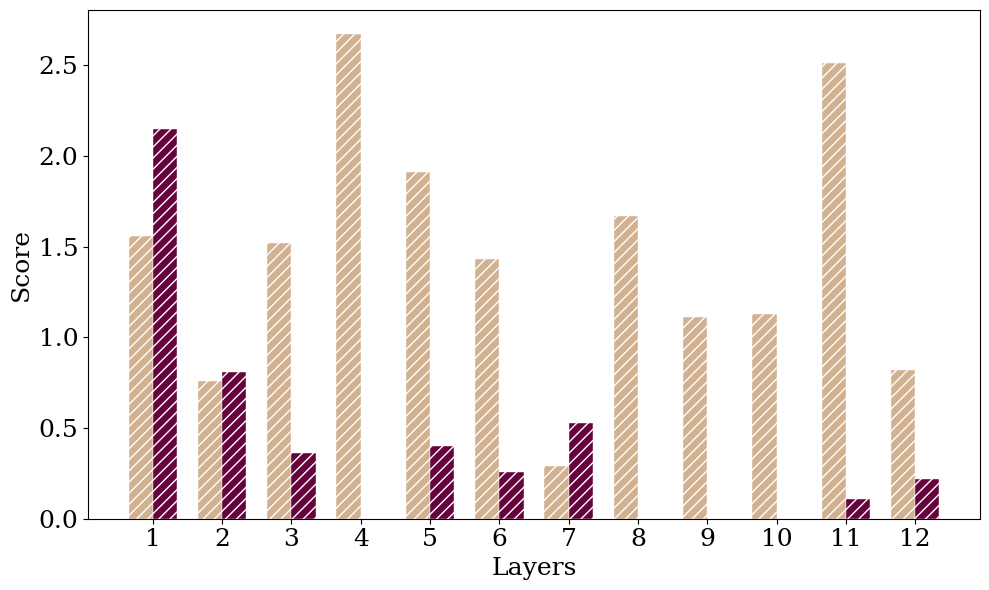}
    \caption{GPT2 on CoQA ($0.97$,$0.68$)}  
    \end{subfigure}
    \begin{subfigure}[b]{0.325\textwidth}
        \includegraphics[width=\linewidth]{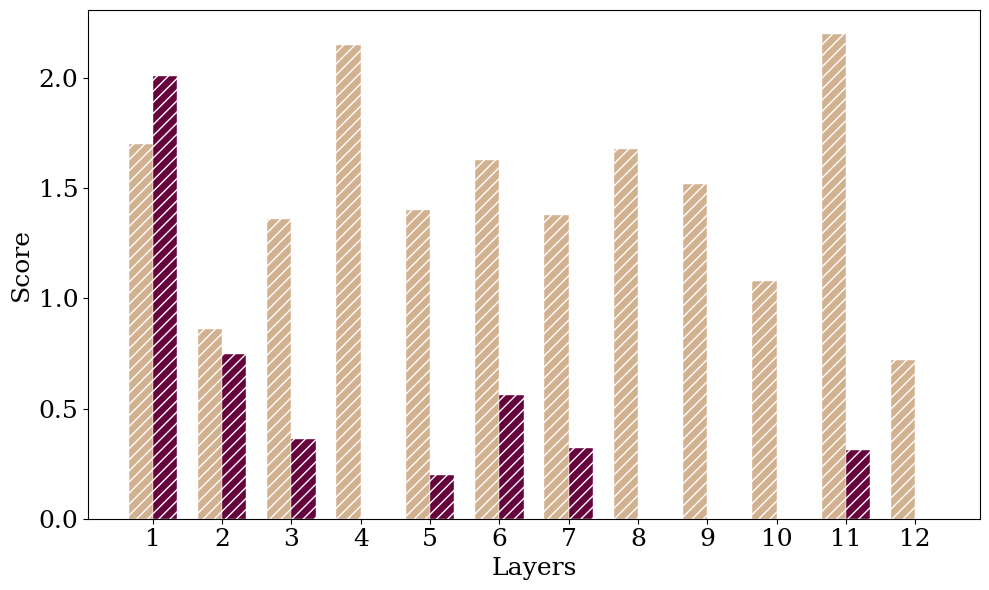}
    \caption{GPT2 on Tatoeba ($0.98$,$0.66$)}  
    \end{subfigure}
\begin{subfigure}[b]{0.325\textwidth}
\includegraphics[width=\linewidth]{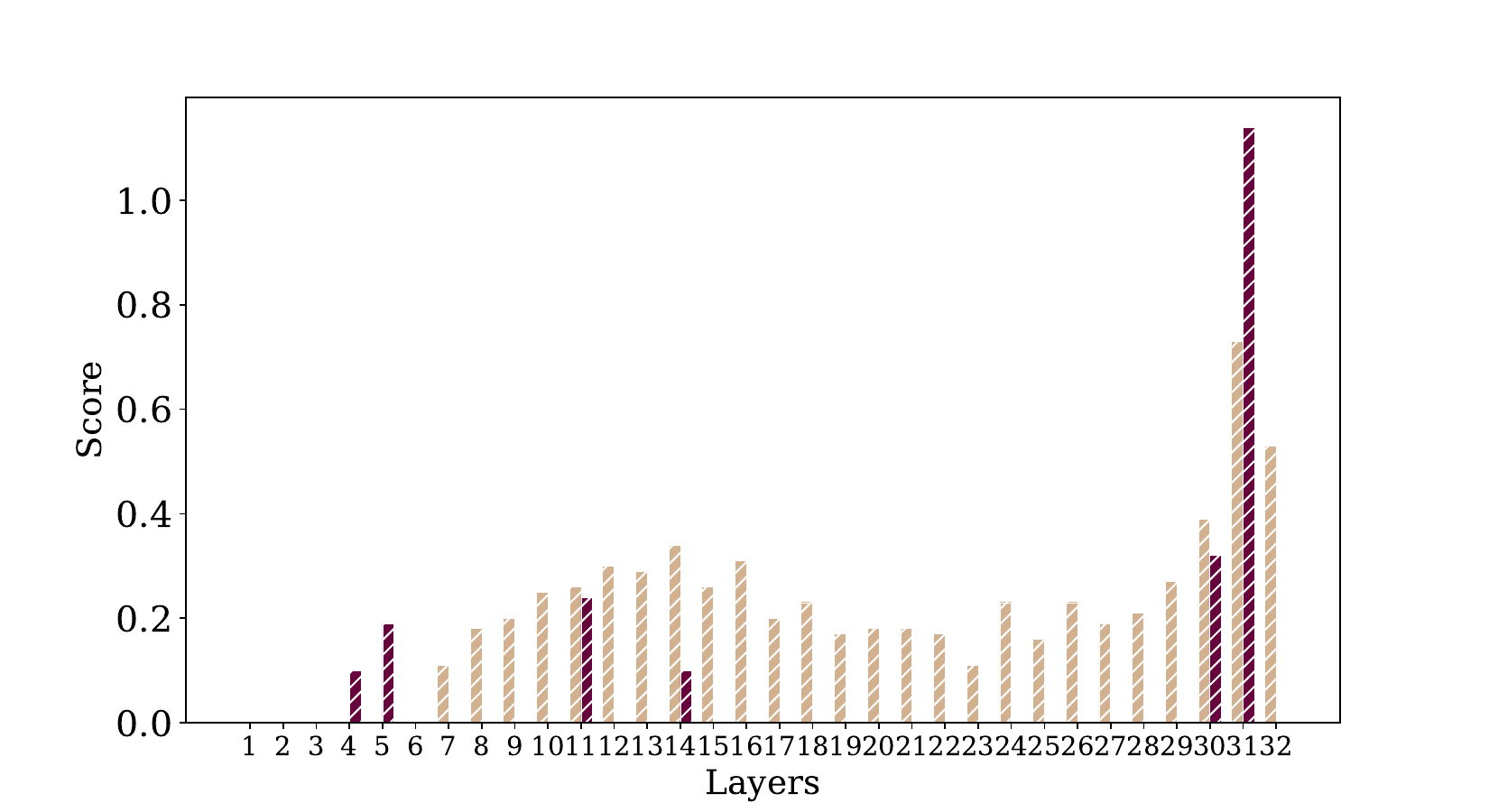}
    \caption{Llama2 on SST-2 ($0.91$, $0.40$)} 
   \end{subfigure}
        \begin{subfigure}[b]
      {0.30\textwidth}
        \includegraphics[width=\linewidth]{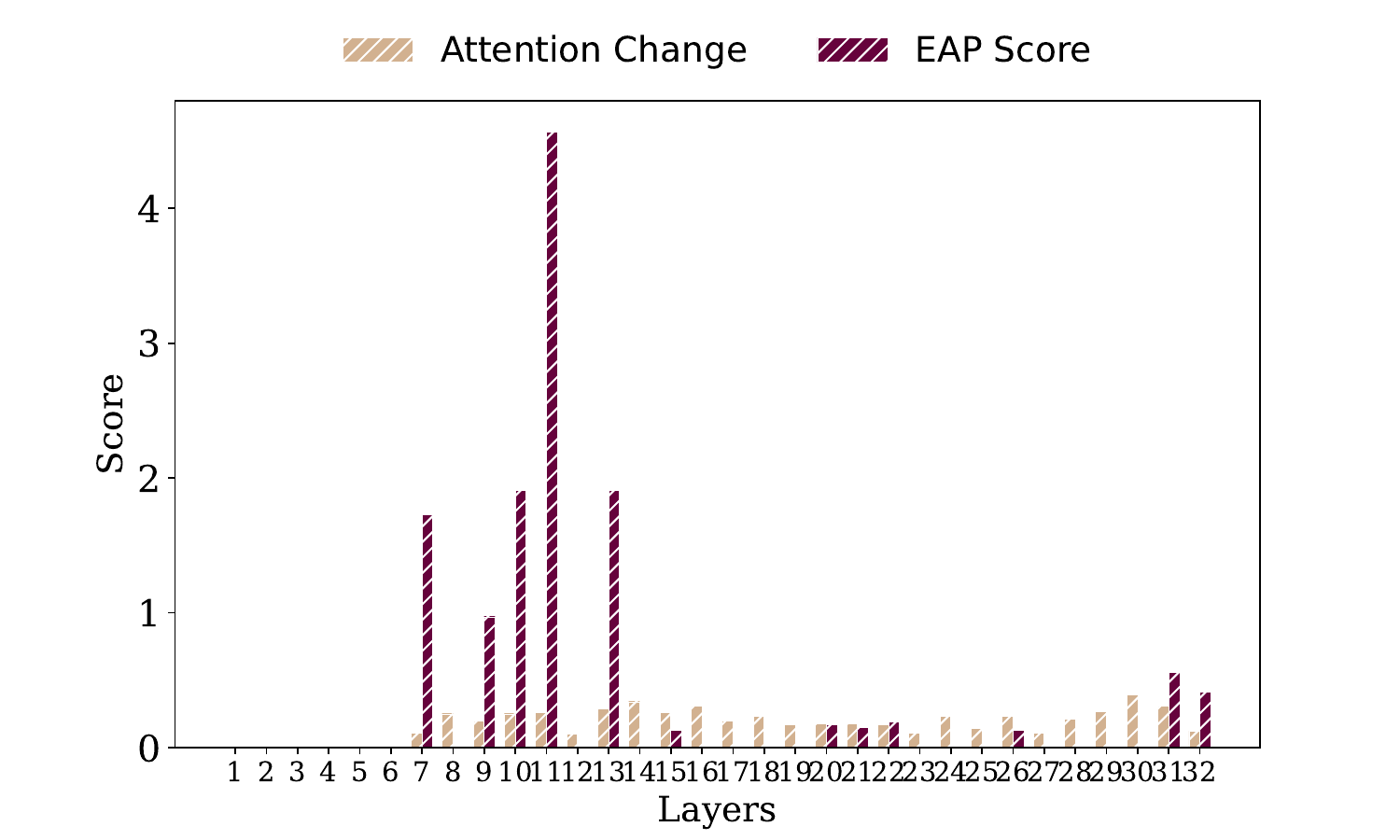}
    \caption{Llama2 on SQuAD ($0.91$, $0.53$)} 
   \end{subfigure}
     \begin{subfigure}[b]{0.35\textwidth}
          \includegraphics[width=\linewidth]{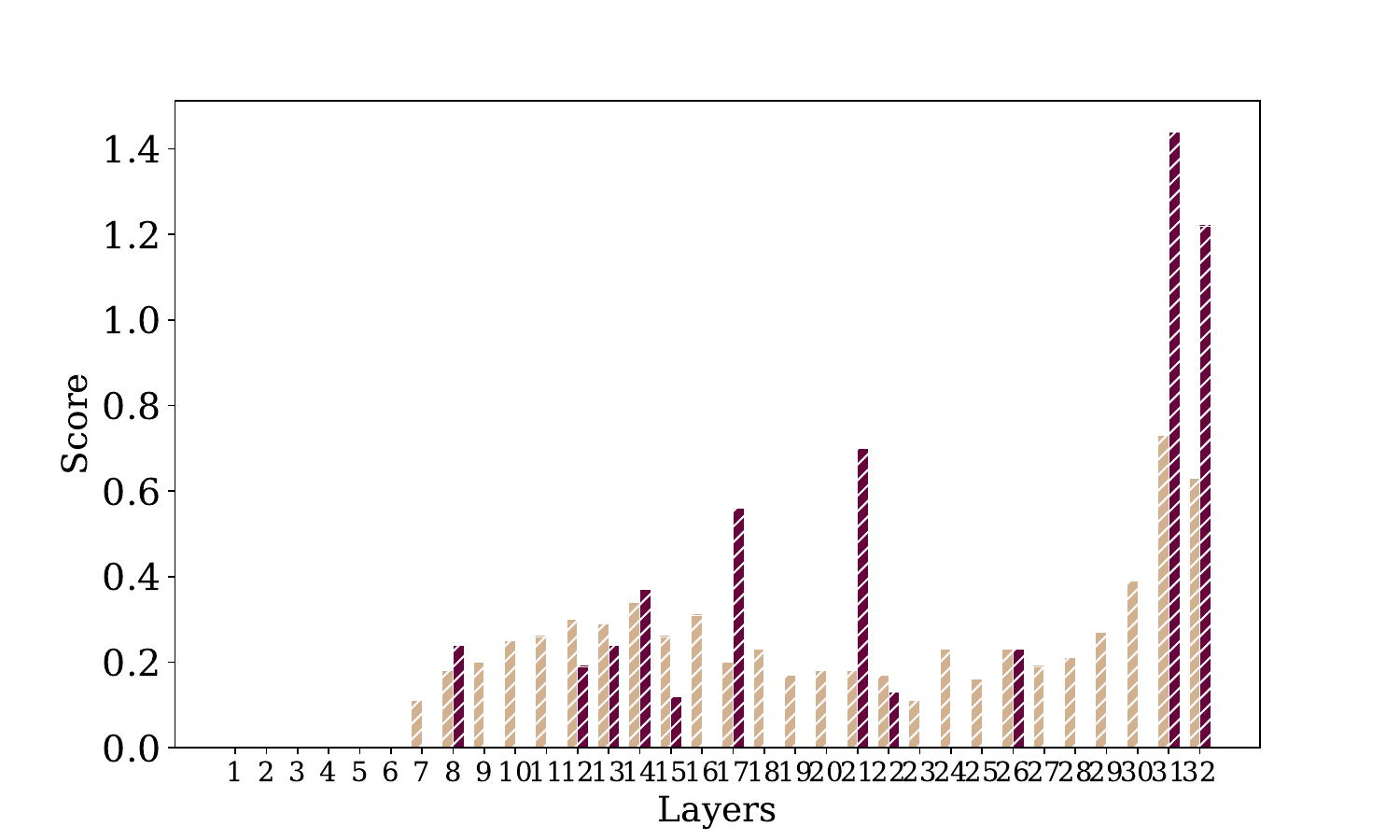}
    \caption{Llama2 on KDE4 ($0.91$, $0.58$)} 
 \end{subfigure}
\begin{subfigure}[b]{0.33\textwidth}
          \includegraphics[width=\linewidth]{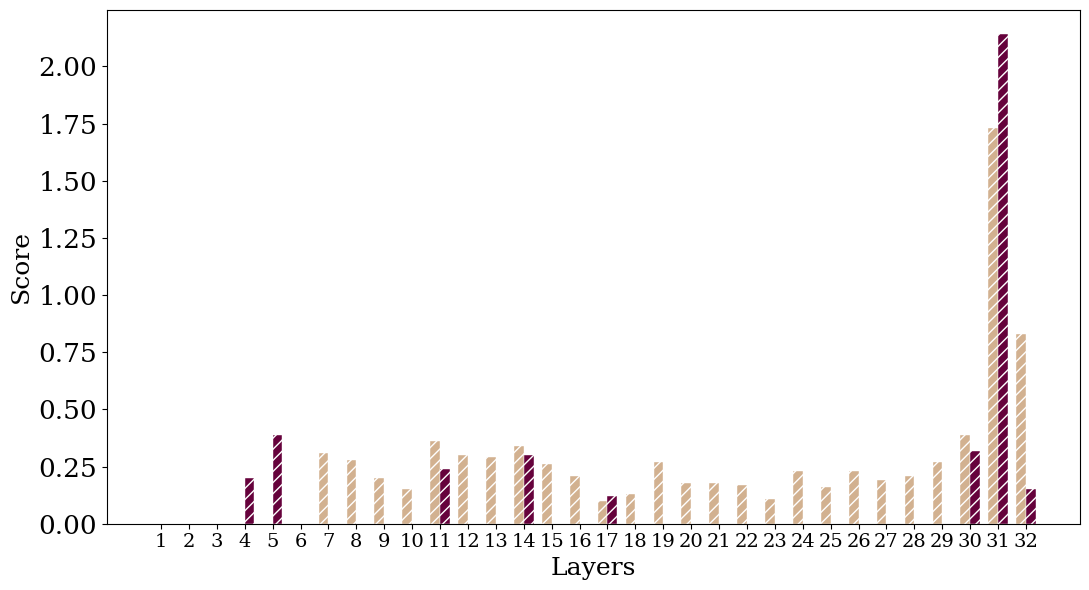}
    \caption{Llama2 on Yelp ($0.92$,$0.59$)} 
    
   \end{subfigure}
            \begin{subfigure}[b]{0.325\textwidth}
          \includegraphics[width=\linewidth]{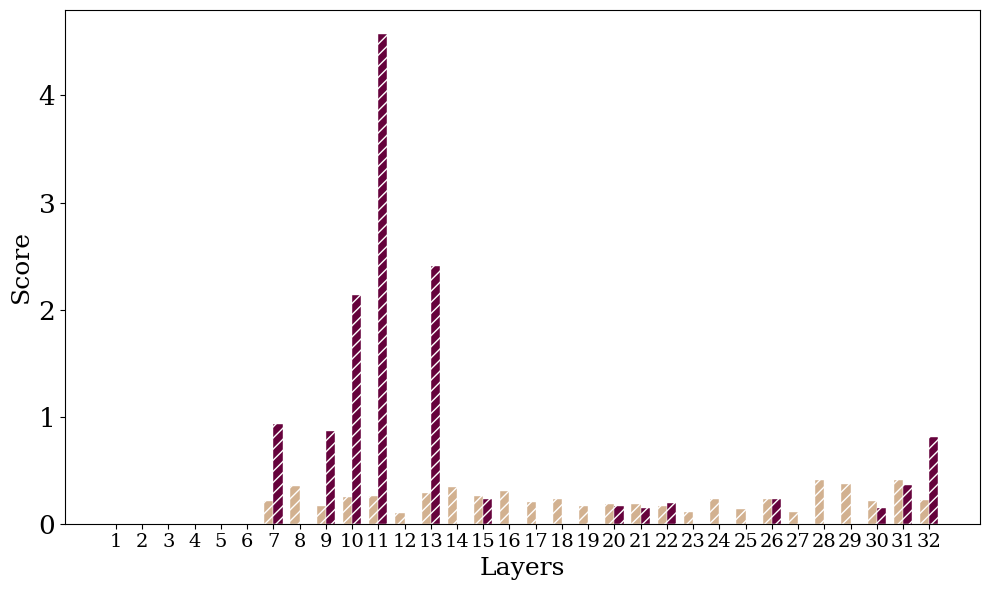}
    \caption{Llama2 on CoQA ($0.97$,$0.57$)} 
    
   \end{subfigure}
            \begin{subfigure}[b]{0.30\textwidth}
          \includegraphics[width=\linewidth]{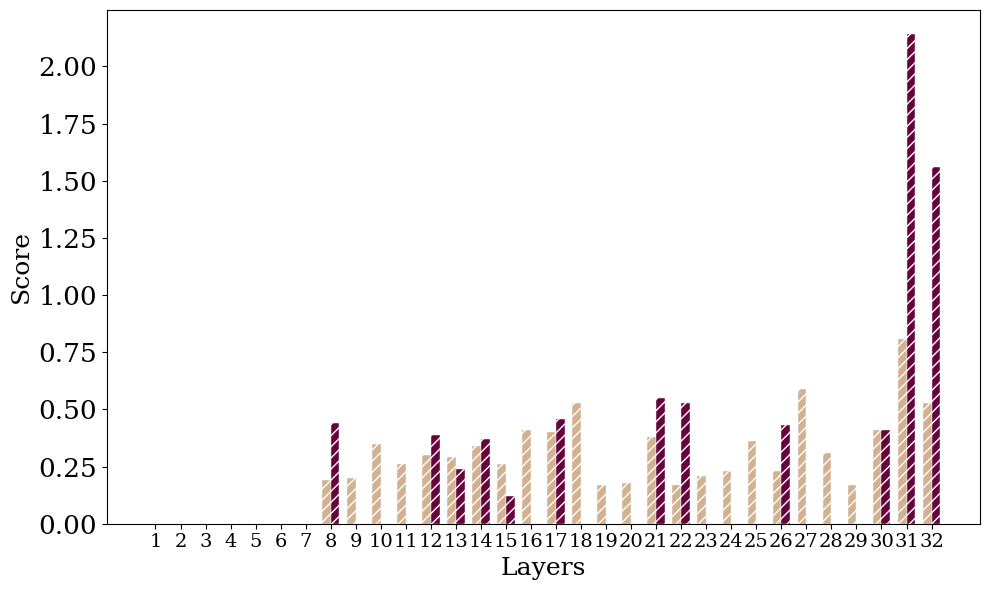}
    \caption{Llama2 on Tatoeba ($0.90$,$0.63$) } 
    
   \end{subfigure}
\caption{Layer-wise distribution of average attention KL divergence and EAP scores for GPT-2 Small and Llama-2-7B across Sentiment Classification, Question Answering, and Machine Translation. }
  \label{fig:gpt2-attn-eap}
\end{figure*}

\begin{figure*}
  \begin{subfigure}[b]{0.30\textwidth}
          \includegraphics[width=\linewidth]{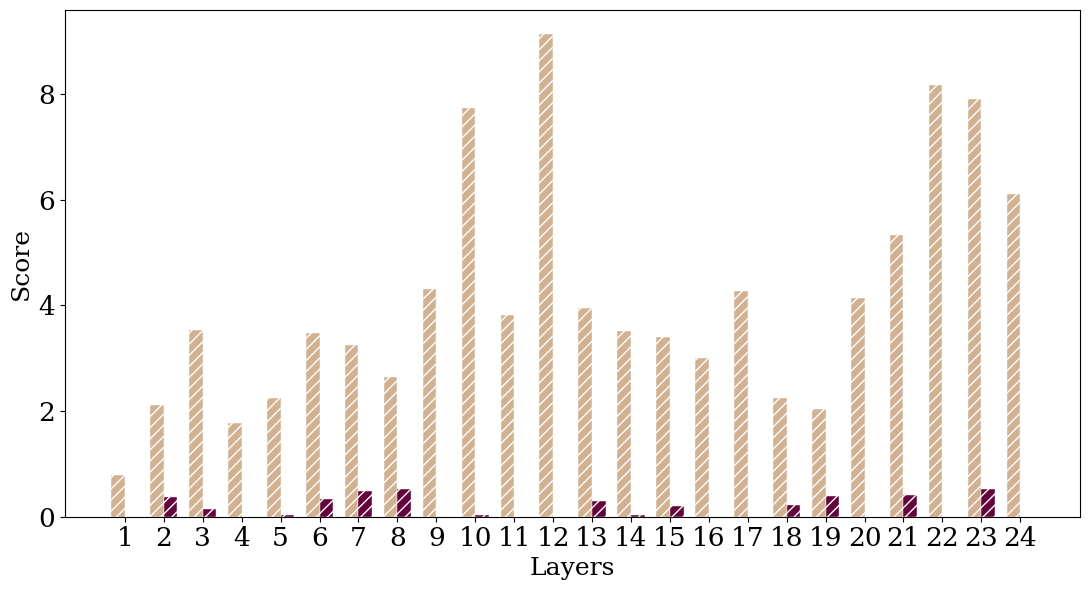}
    \caption{Qwen on Yelp ($0.96$, $0.51$) } 
   \end{subfigure}
     \begin{subfigure}[b]{0.30\textwidth}
          \includegraphics[width=\linewidth]{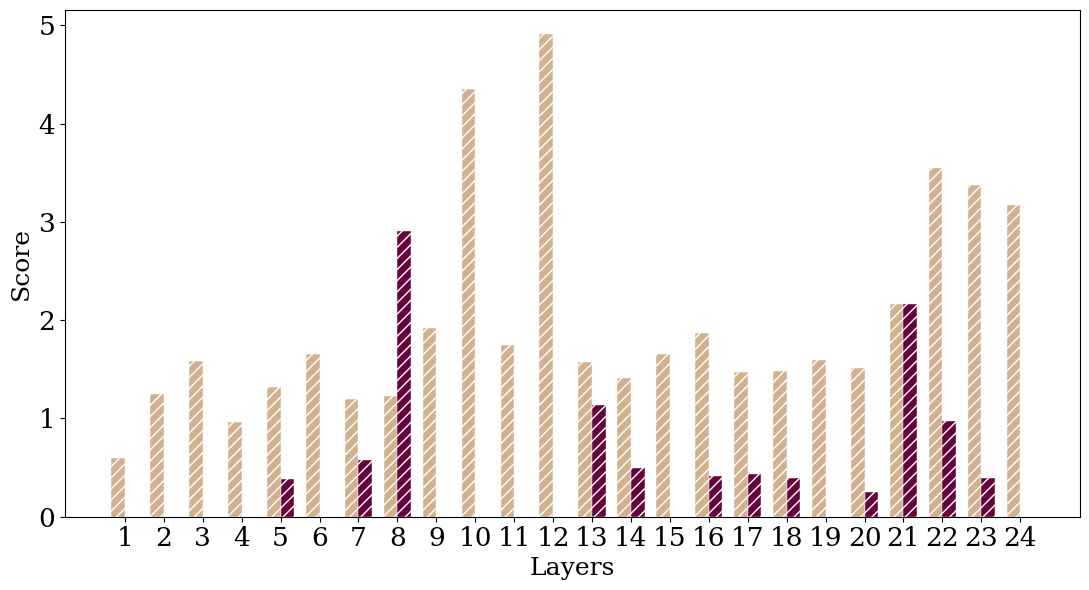}
    \caption{Qwen on CoQA ($0.95$,$0.72$)} 
   \end{subfigure}
        \begin{subfigure}[b]{0.30\textwidth}
          \includegraphics[width=\linewidth]{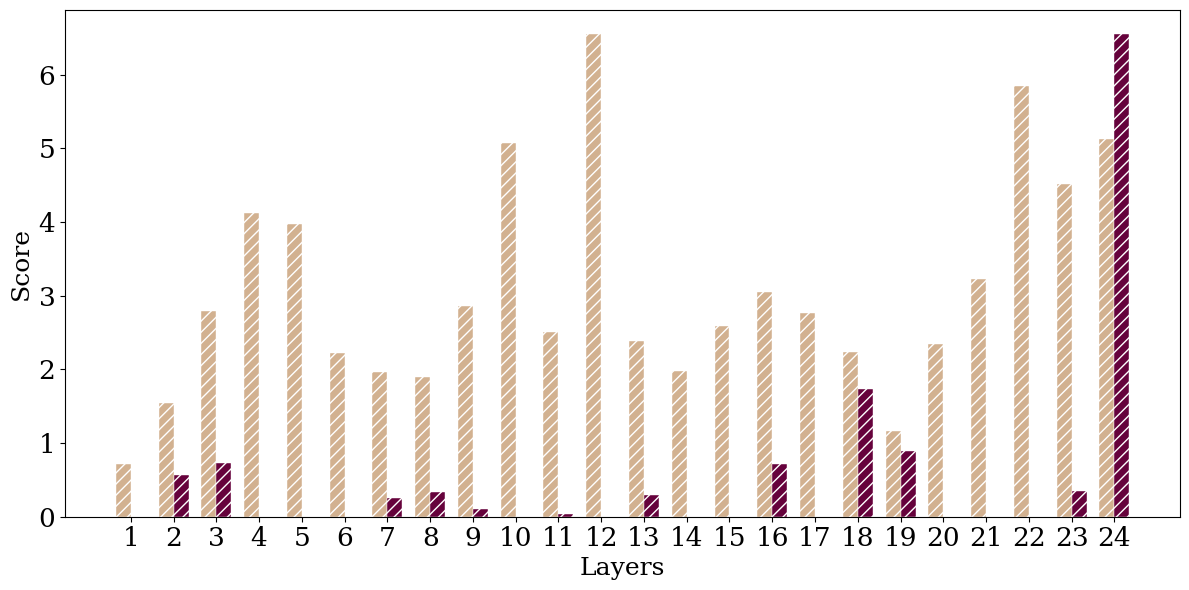}
    \caption{Qwen on Tatoeba ($0.96$,$0.68$)} 
   \end{subfigure}
      \begin{subfigure}[b]{0.34\textwidth}
          \includegraphics[width=\linewidth]{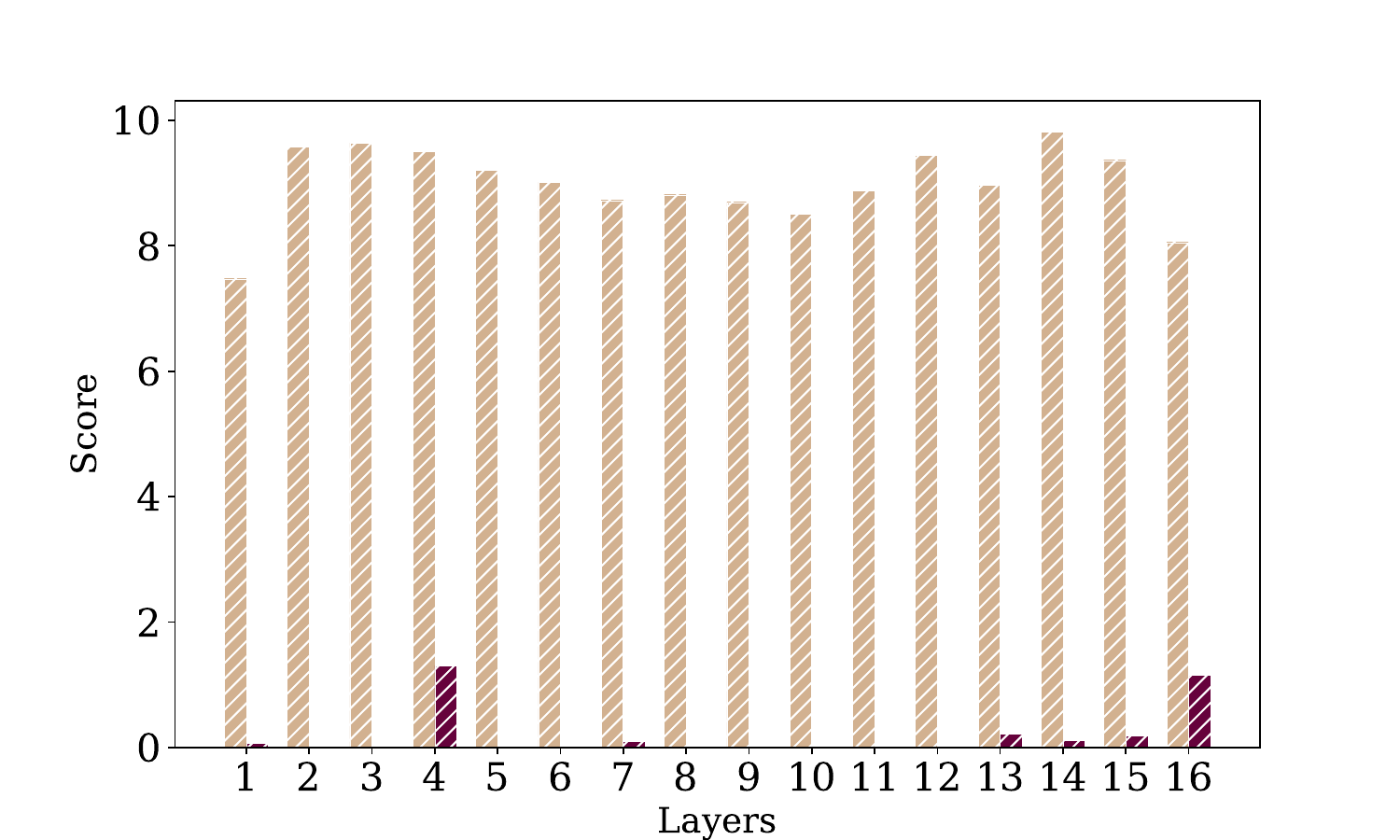}
    \caption{Llama3.2 on SST-2 ($0.99$, $0.65$)} 
       \end{subfigure}
          \begin{subfigure}[b]{0.3\textwidth}
          \includegraphics[width=\linewidth]{Images/attention_patterns/llama2_yelp.png}
    \caption{Llama3.2 on Yelp ($1.00$,$0.48$)} 
   \end{subfigure}
            \begin{subfigure}[b]{0.33\textwidth}        \includegraphics[width=\linewidth]{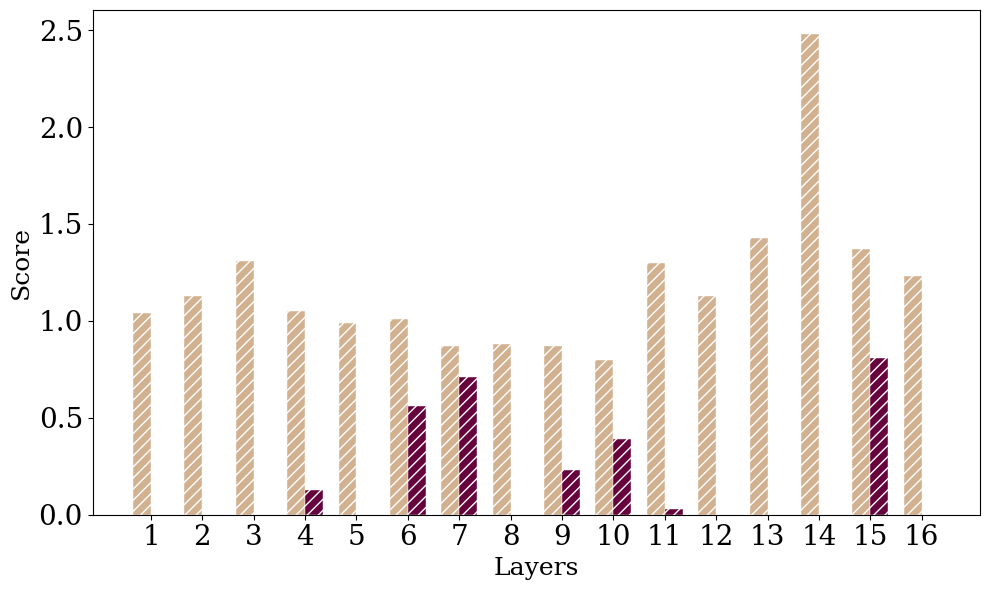}
    \caption{Llama3.2 on Tatoeba ($0.98$,$0.61$)} 
   \end{subfigure}
\caption{Layer-wise distribution of average attention KL divergence and EAP scores for Qwen2 and Llama-3.2-1B across Sentiment Classification (Yelp), Question Answering (CoQA), and Machine Translation (Tatoeba). Counterpart of \autoref{fig:eap_attention} .}
  \label{fig:attn-eap}
\end{figure*}

\begin{figure*}[t]
  \centering
\includegraphics[width=\textwidth]{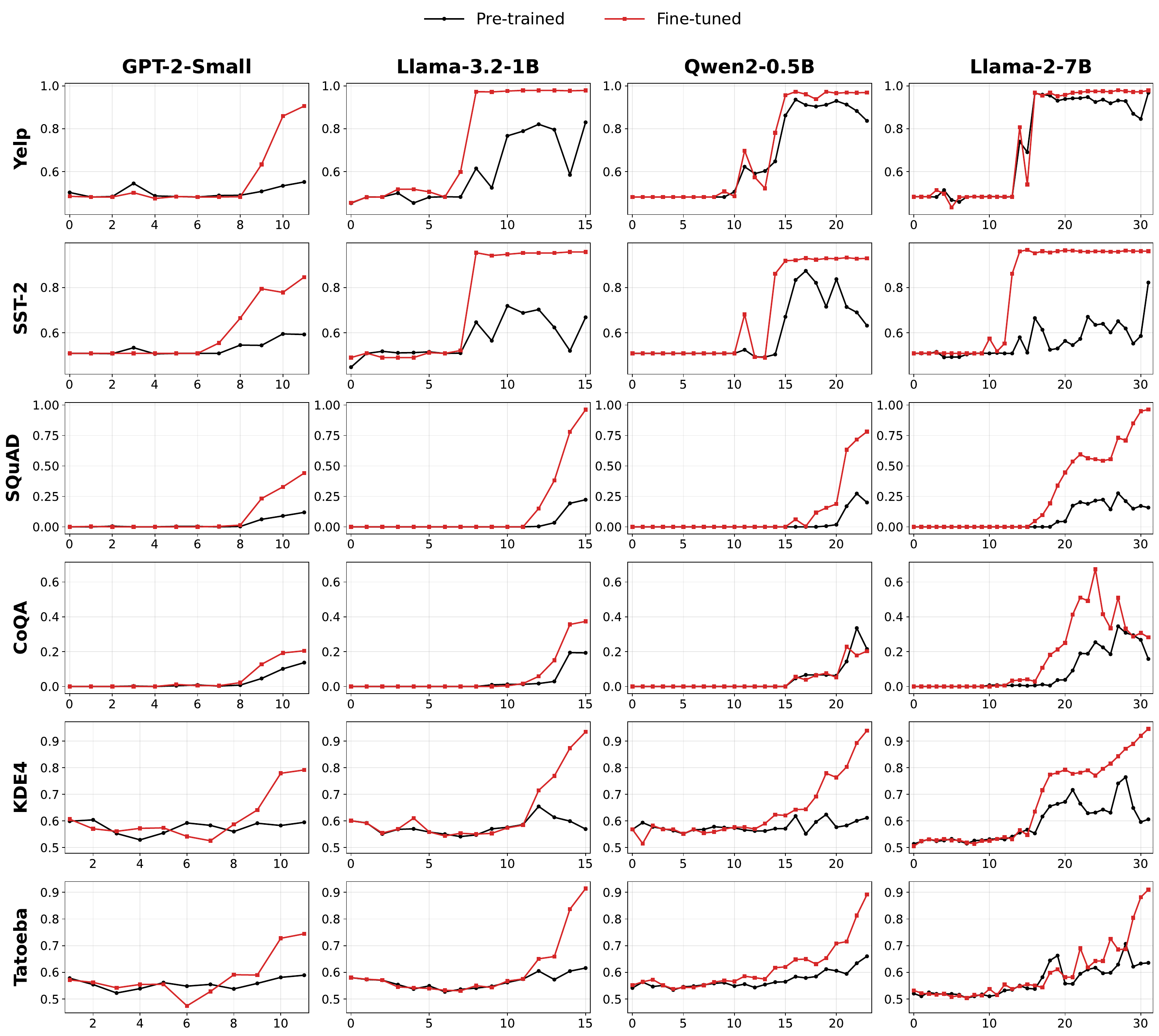}
  \caption{Layer-wise probing of pre-trained vs.\ fine-tuned models, full version of
      \autoref{fig:probing}. \textbf{Rows} are datasets (Yelp, SST-2, SQuAD, CoQA, KDE4, Tatoeba); \textbf{columns} are models.(GPT-2 Small, Llama-3.2-1B, Qwen2-0.5B, Llama-2-7B). Higher = more task-relevant information is already recoverable at that layer. (X-axis: layer index; Y-axis: binary accuracy for SST-2, SQuAD F1 Score for SQuAD, BERTScore for KDE4)}
  \label{fig:gpt2-probing}
\end{figure*}

This section contains the extended results of \autoref{fig:eap_attention} (see \autoref{fig:gpt2-attn-eap} \autoref{fig:attn-eap}),
\autoref{fig:probing} (see \autoref{fig:gpt2-probing}), and \autoref{fig:performance_overlap} (see \autoref{fig:gpt2-correlation}) for rest of models and datasets, which exhibit the same trends and confirm that our findings hold across all four model families.

\end{document}